\documentclass{article}

\usepackage[a4paper, margin=1in]{geometry}
\usepackage[utf8]{inputenc}
\usepackage[round]{natbib}
\usepackage[hidelinks]{hyperref}
\usepackage[font=small,labelfont=bf]{caption}
\usepackage{amsfonts}
\usepackage{authblk}

\usepackage[parfill]{parskip}
\usepackage[title,titletoc]{appendix}

\usepackage{booktabs}
\usepackage{multirow}
\usepackage{graphicx,graphbox,wrapfig,lipsum}
\usepackage{amsmath}
\usepackage{mathtools}

\DeclareMathOperator*{\argmin}{arg\,min}

\usepackage[dvipsnames]{xcolor}
\definecolor{myGreen}{rgb}{0,0.57,0}
\definecolor{myBlue}{rgb}{0,0.39,0.65}

\def\email#1{{\tt#1}}
\usepackage{marvosym} 

\providecommand{\keywords}[1]
{
  \small	
  \textbf{Keywords:} #1
}

\title{Disentangling representations of retinal images with \\ generative models}

\author[1,2 (\Letter)]{Sarah Müller}
\author[1,2,3]{Lisa M. Koch}
\author[4]{Hendrik P. A. Lensch}
\author[1,2 (\Letter)]{Philipp Berens}

\affil[1]{Hertie Institute for AI in Brain Health, Faculty of Medicine, University of Tübingen}
\affil[2]{Tübingen AI Center, Tübingen, Germany}
\affil[3]{Department of Diabetes, Endocrinology, Nutritional Medicine and Metabolism UDEM, Inselspital, Bern University Hospital, University of Bern, Bern, Switzerland}
\affil[4]{Department of Computer Science, University of Tübingen, Tübingen, Germany}

\affil[ ]{\email{\{sar.mueller,philipp.berens\}@uni-tuebingen.de}}

\begin{document}

\maketitle

\begin{abstract}
Retinal fundus images play a crucial role in the early detection of eye diseases. However, the impact of technical factors on these images can pose challenges for reliable AI applications in ophthalmology. For example, large fundus cohorts are often confounded by factors such as camera type, bearing the risk of learning shortcuts rather than the causal relationships behind the image generation process. Here, we introduce a population model for retinal fundus images that effectively disentangles patient attributes from camera effects, enabling controllable and highly realistic image generation. To achieve this, we propose a disentanglement loss based on distance correlation. Through qualitative and quantitative analyses, we show that our models encode desired information in disentangled subspaces and enable controllable image generation based on the learned subspaces, demonstrating the effectiveness of our disentanglement loss. The project code is publicly available\footnote{\url{https://github.com/berenslab/disentangling-retinal-images}}.
\end{abstract}

\section{Introduction}

\begin{figure}[htbp]
    \centering
    \includegraphics[width=\textwidth]{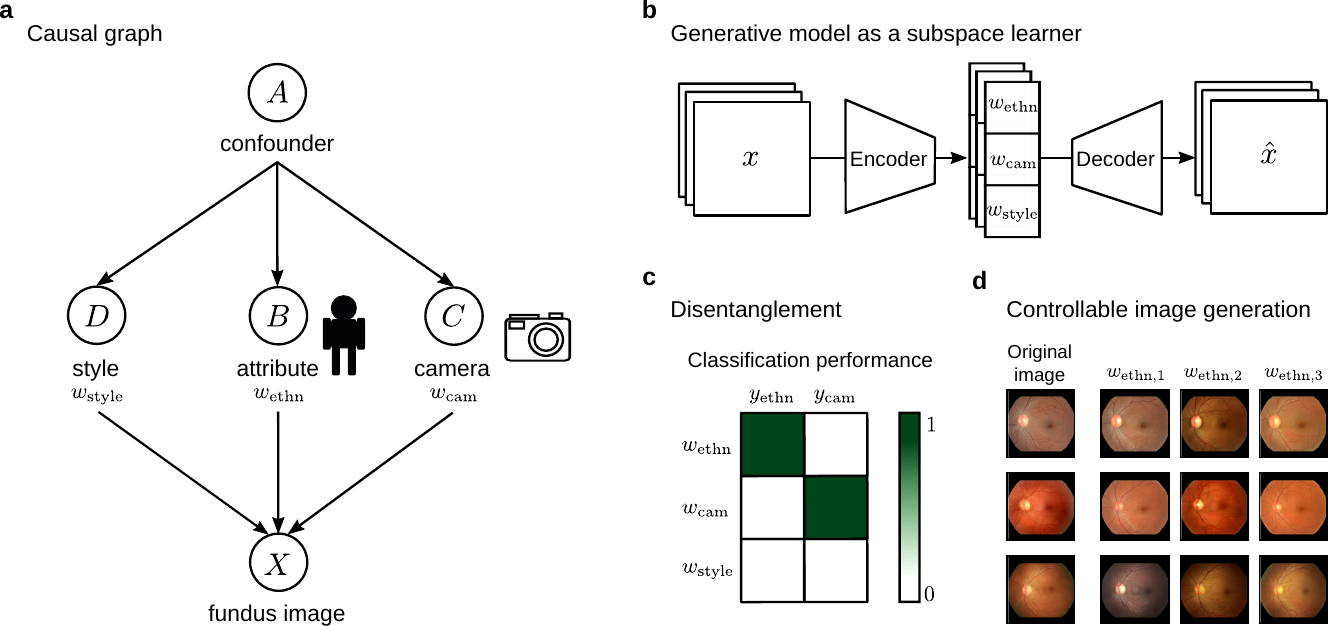}
    \caption{\textbf{Disentangling representations of retinal images with generative models.} In panel \textbf{a}, we present our conceptual causal graph outlining the image generation process. Within this framework, we break down retinal fundus image generation into the causal variables patient attribute (B), camera (C), and style (D) and highlight an unknown confounder (A), which can lead to shortcut connections for deep learning models. In panel \textbf{b}, we introduce a strategy to prevent shortcut learning by using a generative model as an independent subspace learner. Panel \textbf{c} shows our metric for disentanglement: an ideal confusion matrix between the available labels (columns) and the learned subspaces (rows). Panel \textbf{d} shows controllable fundus image generation from swapped ethnicity subspaces.}
    \label{fig:intro}
\end{figure} 

Retinal fundus images are medical images that capture the back of the eye and show the retina, optic disc, macula, and blood vessels. Due to their nature as color photographs acquired through the pupil, they are inexpensive, non-invasive, and quick to obtain, making them available even in resource-constrained regions. Despite their high availability, fundus images can be used to detect the presence of not only eye diseases, but also cardiovascular risk factors or neurological disorders using deep learning \citep{poplin2018pred, kim2020effects, rim2020prediction, son2020predicting, sabanayagam2020deep, xiao2021screen, rim2021deep, nusinovici2022retinal, cheung2022deep, tseng2023validation}.

However, deep learning models require large datasets, and large medical imaging cohorts are often heterogeneous, with technical confounding being ubiquitous. For fundus images, variations in camera types, pupil dilation, image quality, and illumination levels can affect image generation. Additionally, biases due to different hospital sites and patient study selection can lead to spurious feature correlations of these factors with biological variations in the data. Deep learning models run the risk of learning shortcuts based on such spurious correlations \citep{geirhos2020deep}, instead of specific patient-related image features (Fig.\,\ref{fig:intro}\,\textbf{a}). As a result, they may only perform well within the same distribution as training data and latent space analyses can be misleading \citep{mueller2022a}. For example, a retinal imaging dataset could be collected from two hospitals using cameras from different manufacturers \citep{fay2023avoid}. Camera A produces images that appear different in hue from camera B. The first hospital may treat predominantly Latin American patients, while the second hospital may treat predominantly Caucasians. In this scenario, it is simpler for a deep learning model to infer a patient's ethnicity from a fundus image by recognizing the hue rather than understanding the hidden causal relationship between ethnicity and phenotypic image features \citep{castro2020causality}.

One approach to address data confounders is subspace learning, which combines representation learning with disentanglement. In our work, we constructed latent subspaces based on a causal graph, leveraging domain knowledge about the factors influencing the image generation process \citep{castro2020causality}. We used a simplified causal model that divides fundus image generation into three causal variables: patient attribute (B), camera (C), and style (D) (Fig.\,\ref{fig:intro}\,\textbf{a}). Style features in the literature of causal representation learning and image generation are attributed to information which is agnostic to content features (B and C in our case) \citep{yao2024multiview,Havaei2021cond}. Due to the presence of an unknown confounder (A), we assume that there is a statistical association between B, C, and D, even though they are not directly causally related. This association could lead to shortcut connections driven by spurious correlations, particularly between factors like patient attributes (e.g., ethnicity) and technical factors like the camera (Fig.\,\ref{fig:intro}\,\textbf{a}). To prevent shortcut learning, we disentangle these factors of variation by encoding them into statistically independent latent subspaces \citep{bengio2013rep, higgins2018}.

One common strategy for acquiring disentangled representations is to use generative image models such as variational autoencoders (VAEs) \citep{kingma2014autoencoding}. Generative models offer a valuable inductive bias for representation learning by reconstructing images from the latent space and can be extended to learn independent subspaces (Fig.\,\ref{fig:intro}\,\textbf{b}) \citep{higgins2017betavae, klys2018learning, tschannen2018recent}. Previous methods primarily focused on achieving a disentangled latent space, often neglecting the simultaneous pursuit of high-quality image generation. However, when latent space disentanglement and image generation are addressed simultaneously, generative image models provide a built-in interpretation method through controllable feature changes in the image space (Fig.\,\ref{fig:intro}\,\textbf{d}).

In this work, we close the gap of missing generative disentangled representation learning in conjunction with high-quality image generation in the domain of fundus images. Our contributions are threefold
\begin{itemize}
    \item[1.] Disentangling latent subspaces in a state-of-the-art generative adversarial model with a disentanglement loss based on distance correlation.
    \item[2.] Developing a method that jointly optimizes for disentanglement and controllable high-resolution image generation end-to-end.
    \item[3.] Conducting extensive analysis on the effectiveness of our disentanglement loss using both quantitative and qualitative measures to evaluate disentanglement, image quality, and controllable image generation.
\end{itemize}

\section{Causal background}
Deep generative models have demonstrated great success in data generation and density estimation from observational data, but they face limitations such as lack of explainability, risk of learning spurious correlations and poor performance on out-of distribution data. To address these challenges, causal models offer several valuable benefits, such as distribution shift robustness, fairness, and interpretability \citep{scholkopf2021toward}. The interpretability of causal models can be viewed in terms of both learning semantically meaningful representations of a system and controllably generating realistic counterfactual data. Causal models, particularly Structural Causal Models (SCMs), describe the data-generating processes and the complex causal relationships between variables, making them a natural complement to generative models \citep{scholkopf2021toward}. Causal representation learning (CRL) focuses on identifying meaningful latent variables and their causal structures from high-dimensional data, ensuring the correct identification of the true generative factors \citep{scholkopf2021toward}.

In our causal diagram the fundus image (X) is influenced by several factors: patient attribute (B), camera (C), and style (D). However, there is also an unknown confounder (A) that affects all of these factors and creates a statistical association between the causal variables B, C, D, even though they are not causally related \citep{castro2020causality,scholkopf2021toward}. We represent the causal relationships between these variables by a directed acyclic graph (DAG) which implies the factorization of causal conditionals \citep{scholkopf2021toward}:
\begin{align*}
    P(X) = P(B|A) \cdot P(C|A) \cdot P(D|A).
\end{align*}
In theory, the confounder could be addressed by causal interventions, which is not feasible in our setting with only observational data. Therefore, in this work, we address the confounder by disentangling the causal variables B, C, D in the latent space of a generative model. Our approach is inspired by our DAG structure, which implies a disentangled factorization of causal conditionals due to the causal Markov condition \citep{castro2020causality}. Regarding CRL, our causal variables B and C are identifiable because of the invariance principle introduced in \citep{yao2024unifyingc}. For these variables, we use labels together with linear classifiers to enforce content encoding in latent subspaces. Intuitively, with our subspace classifiers we force images with the same label to have the same content latent subspaces. Identifying the style variable (D) is more complicated, but we can use Corollary 3.9-3.11 from \citep{yao2024multiview} to identify style, assuming that it is independent of the content variables B and C.

\section{Related work} \label{sec:related_work}

\paragraph{Generative models for representation learning.}
The most commonly used generative models for representation learning are VAEs \citep{tschannen2018recent}. However, VAEs produce blurry images that lack details, which are relevant in the field of retinal imaging, where fine details have a major impact on patient attribute or disease classification. Consequently, work on generative representation learning with VAEs focuses primarily on learning low-dimensional latent spaces with specific properties, rather than on generating accurate image reconstructions \citep{tschannen2018recent}. 

Generative adversarial networks (GANs), on the other hand, succeed in generating fine details in the image domain and can be adapted to representation learning by GAN-inversion \citep{gosh2022inv}. Especially state-of-the art GANs like StyleGAN2 \citep{karras2020analyzing} are able to create high-fidelity and high-resolution images. When labels are available, conditional GANs \citep{mirza2014conditional} and their variants \citep{odena2017conditional, miyato2018cgans} are the most established extensions for class-conditional image generation. However, conditional GANs mainly focus on generating realistic images and do not focus on the statistical independence between the latent space and the conditioned label. 

InfoGAN \citep{chen2016infogan,paul2021unsupervised} modifies the GAN objective to learn unsupervised disentangled representations by maximizing a lower bound of the mutual information (MI) between a subset of the latent variables and the observation. However, this extension does not guarantee that for complex image data their induced factored latent codes are not dominated by shortcut features such as camera type. Furthermore, due to the unsupervised setting, disentangled features are analyzed post-hoc with human feedback, leading to potential non-identifiability \citep{locatello2019ch}, especially in highly specialized domains such as medical imaging. Semi-StyleGAN \citep{nie2020semisupervised}, closest to our work, is a special case of semi-supervised conditional GANs and extends the StyleGAN architecture to an InfoStyleGAN with a mutual information loss and addresses non-identifiability by adding a weakly supervised reconstruction term for partially available factors of variation. However, they do not address the problem of potential spurious correlations in the data and the resulting need for statistically independent latent factors.

Diffusion probabilistic models (DPMs) show remarkable success in image generation. However, since the latent variables of DPMs lack high-level semantic meaning, little attention has been paid to representation learning with DPMs. There are few works that combine disentangled representation learning with DPMs. In \citep{preechakul2021diffusion,zhang2022unsupervised}, they explore DPMs for representation learning via autoencoding by jointly training an encoder for high-level semantics and a conditional DPM as a decoder for image reconstruction. In \citep{yang2023disdiff}, they eliminate semantic linear classifiers and achieve and unsupervised disentanglement by decomposing the gradient fields of DPMs into sub-gradient fields for different generative factors. However, compared to the StyleGAN2 architecture used in this work, DPMs lack the ability to control scale-specific image generation, the learned representations may not be easy to interpret, and image generation speed is considerably slower.

\paragraph{Dependence estimators for disentanglement.}
In the field of disentanglement, a common goal is to use generative models as independent subspace learners, necessitating the estimation and minimization of statistical dependence between subspaces. Mutual information (MI) is a viable measure for quantifying the mutual dependence between two variables, effectively capturing nonlinear dependencies. However, estimating MI in high-dimensional spaces remains a challenge. Therefore, recent work focused on estimating bounds to establish tractable and scalable MI objectives. Most of the proposed approaches are lower bounds \citep{poole2019on, belghazi2018mut}, which are not applicable to MI minimization problems. There is limited work on estimating MI upper bounds, and most of them are restrictive in the sense that they require the conditional distribution to be known \citep{alemi2017deep, poole2019on, cheng2020club}.

Distance measures between distributions, such as the MI, provide a theoretically sound measure, but often require density estimation. To address this, non-parametric kernel method such as maximum mean discrepancy (MMD) \citep{sejdinovic2013eq} offer a principled alternative. MMD simplifies distance computation by using mean kernel embeddings of features, and is therefore also used to learn invariant representations, as in the Variational Fair Autoencoder (VFAE) \citep{louizos2017var}. Another non-parametric alternative to MI estimation are adversarial classifiers, which solve the invariant subspace problem by solving the trade-off between task performance and invariance through iterative minimax optimization \citep{ganin2015unsup, ganin2016domain, xie2017contr}. However, adversarial training adds an unnecessary layer of complexity to the learning problem \citep{moyer2018inv} that does not scale well with multiple subspaces. In this work, we use distance correlation \citep{sz2007dCor}, as another non-parametric measure that does not require any pre-defined kernels. Distance correlation is fast to evaluate and captures the linear and nonlinear dependence between two random vectors of arbitrary dimensions. It proves to be of versatile use  in deep learning, ranging from measuring disentanglement to measuring similarity between networks \citep{liu2020metrics,zhen2022versatile}.

\paragraph{Disentanglement in medical image analysis.}
In medical imaging, disentangled representations are relevant to many research questions related to image synthesis, segmentation, registration, and causal- or federated learning, e.g., for disease decomposition, harmonization, or controllable synthesis \citet{liu2022learn}. For example, \citet{fay2023avoid} present a predictive approach to avoid shortcut learning for MRI data from different scanners. To minimize the dependencies between a task-specific factor (e.g. patient attribute) and a confounding correlated factor (e.g. scanner), they estimate a lower bound for MI. However, as mentioned above, a lower bound MI estimate is inconsistent with MI minimization tasks. In ICAM \citep{bass2020icam, bass2021icam}, the authors build on a VAE-GAN approach for image-to-image translation \citep{lee2018diverse} to disentangle disease-relevant from disease-irrelevant features. However, disentanglement in ICAM requires longitudinal data, where patient images are available at different disease stages. In a similar approach, \cite{ouyang2021representation} embed MRI images into independent anatomical and modality subspaces by image-to-image translation. Like ICAM, they have the data constraint that their method only works with images of the same patient from different modalities. In another attempt to explain classifier features, StylEx \citep{lang2021explaining} integrates a classifier into StyleGAN training. They have a classifier-specific intermediate latent space and can visualize image features that are important for the classifier decision. Although the authors apply their approach to the retinal fundus domain, they do not address potential spurious correlations with latent space disentanglement. In \cite{galati2023A2V}, another work based on StyleGAN, they present a semi-supervised adaptation framework for brain vessel segmentation between two imaging domains: magnetic resonance angiography to venography. However, here they assume that all vessel properties are disentangled within the StyleGAN latent space and do not add any new regularizers regarding latent space disentanglement. In \cite{Havaei2021cond} they train a conditional adversarial generative model with two latent spaces that encode style and content which they also disentangle with two regularization steps. However, compared to our method, they only generate low resolution images and do not focus on the generation of high frequency details. In addition, their optimization procedure is very involved compared to ours, with two generators and five discriminators.

\section{Methods}
This section introduces the proposed model extensions and training constraints to disentangle representations, prevent shortcut learning, and still enable controllable high-resolution image generation. First, we introduce the dataset, an encoder model and our disentanglement loss as preliminary steps towards the complete image generation pipeline.

\subsection{Data}\label{sec:data}

\begin{figure}[htbp]
    \centering
    \includegraphics[width=\textwidth]{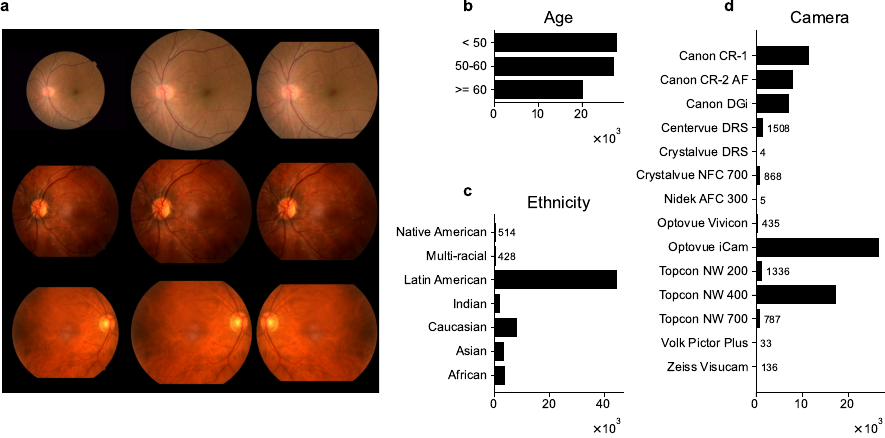}
    \caption{\textbf{Image preprocessing and data distribution.} In panel \textbf{a}, we outline our two-step preprocessing of fundus images. The first column displays original samples from the training set, while the second column shows images tightly cropped around the fundus circle. In the third column, we further cropped and flipped the images for consistency. The corresponding label distributions are visualized as histogram plots in panels \textbf{b-d}. Specifically, panel \textbf{b} illustrates the age group distribution across three classes, panel \textbf{c} shows the ethnicity distribution, and panel \textbf{d} presents the camera distribution.}
    \label{fig:data}
\end{figure}

We trained our models on retinal fundus images provided by EyePACS Inc. \citep{Cuadros2009EyePACSAA,Eyepacs2008Grading}, an adaptable telemedicine system for diabetic retinopathy screening from California. All data was anonymized by the data provider. We filtered for healthy fundus images with no reported eye disease which were labelled as ``good'' or ``excellent'' quality. This resulted in 75,989 macula-centered retinal fundus images of 24,336 individual patients. For our experiments, we split the data by patient-identity into 60\% training, 20\% validation, and 20\% test sets.

The retinal fundus images in EyePACS are not standardized in aspect ratio and field of view (Fig.\,\ref{fig:data}\,\textbf{a}, first column). To do so, we cropped them to a tightly centered circle \citep{mueller2023prep} and resized them to a resolution of $256\times 256$ pixels (Fig.\,\ref{fig:data}\,\textbf{a}, middle column). Furthermore, to address evident and unwanted factors of variation, we standardized the images by masking them to a uniform visible area, as many images had missing regions at the top and bottom. Additionally, we horizontally flipped the images to ensure that all optic discs were positioned on the left side (Fig.\,\ref{fig:data}\,\textbf{a}, last column).

Retinal fundus images from EyePACS come with extensive technical and patient-specific metadata. Here, we relied on age, ethnicity, and camera labels (Fig.\,\ref{fig:data}\,\textbf{b-d}). The camera labels in EyePACS include some duplicate manufacturer names, which we merged into 14 classes (Fig.\,\ref{fig:data}\,\textbf{d}).

\subsection{Encoder model for mapping factors of variation} \label{subsection:encoder model}
We first considered the image encoding task by learning an encoder $f_\theta$ that maps images to a latent representation $f_\theta: \mathcal{X} \rightarrow \mathcal{W}$. We observed noisy image data $x\in\mathcal{X}$ obtained from an unknown image generation process $x=g(s)$, where $s=(s_1, s_2, \dots, s_K)$ are the underlying factors of variation. Our goal was to find a mapping $f_\theta: \mathbb{R}^m \rightarrow \mathbb{R}^d$ for a latent space $f_\theta(x)=w=(w_1, w_2, \dots, w_K)$ such that each factor of variation $s_k$ can be recovered from the corresponding latent subspace $w_k \in \mathbb{R}^{d_k}$, $d=\Sigma_{k=1}^K d_k$, by a linear mapping $\hat{s}_k = C_{\psi_k} w_k$, but not from the other latent subspaces $w_l$ for $l \neq k$.

\begin{figure}[htbp]
    \centering
    \includegraphics[width=0.48\textwidth]{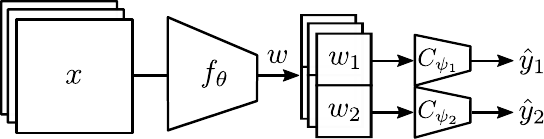}
    \caption{\textbf{Encoder architecture.} Feature encoder $f_\theta$ encodes input images $x$ to feature vectors $w$. Each feature vector is split into two subspaces $w_1$ and $w_2$ by classification heads $C_{\psi_1}$ and $C_{\psi_2}$.}%
    \label{fig:methods_encoder}
\end{figure}

Since we did not know the true underlying factors of variation, we used pseudo-labels from a labeled image dataset $\mathcal{D} = \{(x^{(i)}, y^{(i)})\}_{i=1}^N$, where $y^{(i)}$ is a vector of attribute labels. We assumed that these labels represent, in part, the true underlying factors of variation. We consider discrete attributes $y_k^{(i)} \in \mathbb{N}$, where $y_k^{(i)}$ is the label for the $k^\text{th}$ attribute of the $i^\text{th}$ sample. 

In practice, we worked with retinal fundus images as our input domain $\mathcal{X}$, using a vector of available metadata as labels $y$. The goal was to encode the images into disentangled subspaces: $w=(w_1, w_2)$ for potentially highly correlated labels. To enforce attribute encoding in the subspaces, we defined a predictive model (Fig.\,\ref{fig:methods_encoder}) consisting of two parts: the feature encoder and the classification heads. The feature encoder $f_\theta$ maps input images $x \in \mathbb{R}^{m}$ to a feature vector $w \in \mathbb{R}^{d}$. Each feature vector was split into two subspaces by a linear classification head $C_{\psi_k}$ attached to each subspace $k$. Thus, optimizing the encoder to map to attribute subspaces was equivalent to optimizing the cross-entropy loss for each subspace individually:

\begin{align}
(\theta^*, \psi^*) &= \argmin_{\theta, \psi} \, L_\text{CE}(\theta, \psi),\\
L_\text{CE}(\theta,\psi) &= \frac{1}{K}\sum_{k=1}^K - y_k^T \log C_{\psi_k}(w_k) \label{eq:encoder_loss_classifier}.
\end{align}

\subsection{Disentanglement loss}\label{subsec:disentanglement_loss}
If the predictive model (Fig.\,\ref{fig:methods_encoder}) was only composed of an encoder and subspace classifiers, subspaces could correlate and share information. Therefore, we additionally penalized the presence of shared information with a disentanglement loss by minimizing the distance correlation (dCor) between unique subspace pairs: 

\begin{align}
(\theta^*, \psi^*) &= \argmin_{\theta, \psi} \, L_\text{CE}(\theta, \psi) + \lambda_\text{DC} L_\text{DC}(\theta) \label{eq:encoder_loss},\\
L_\text{DC}(\theta) &= \frac{2}{K(K-1)}\sum_{i=1}^K \sum_{j=1}^{i-1} \text{dCor}(w_i, w_j).\label{eq:encoder_loss_dcor}
\end{align}

Distance correlation \citep{sz2007dCor} measures the dependence between random vectors of arbitrary dimension. It is analogous to Pearson's correlation but can also measure nonlinear dependencies. Moreover, distance correlation is bounded in the $[0, 1]$ range and is zero only if the random vectors are independent.
\paragraph{Examples of different dependence measures.} We provided a few 2D toy examples in Appendix \ref{sec:appendix_toy_example}, where we compared the performance of minimizing distance correlation with minimizing linear dependence measures. When the relationship between the random variables was nonlinear, only minimizing distance correlation showed an effect.

We used the \textit{empirical distance correlation} \citep{sz2007dCor} to measure correlation between batch samples of subspace tensors:
\begin{align}
    \text{dCor}(w_1, w_2) &= \frac{\text{dCov}(w_1, w_2)}{\sqrt{\text{dCov}(w_1, w_1)\,\text{dCov}(w_2, w_2)}}.\label{eq:dcor}
\end{align}
We assumed n-dimensional batches of subspace vectors $w_1 \in \mathbb{R}^{n\times d_1}$ and $w_2 \in \mathbb{R}^{n\times d_2}$. 
Here, dCov is the distance covariance 
\begin{align}
    \text{dCov}^2(w_1, w_2) &= \frac{1}{n^2} \sum_{i=1}^n \sum_{j=1}^n A_{i,j} B_{i,j}. 
\end{align}
Each matrix element $a_{i,j}$ of $A$ is the Euclidean distance between two samples 
\begin{align}
    A \in \mathbb{R}^{n\times n}, \, a_{i,j} &= \|w_1^{(i)} - w_1^{(j)} \|_2, \,i,j = 1,2,\dots, n
\end{align}
after subtracting the mean of row $i$ and column $j$, as well as the matrix mean. B is similarly calculated for $w_2$.

\subsection{Generative image model} \label{subsection:generative model}

\begin{figure}[htbp]
    \centering
    \includegraphics[width=\textwidth]{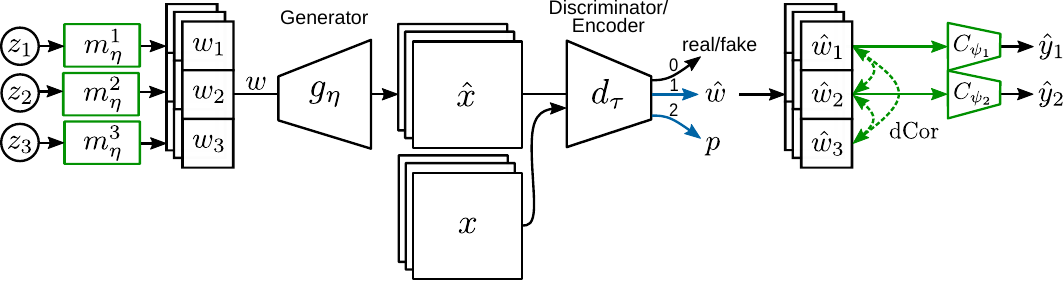}
    \caption{\textbf{Generative image model with subspace constraints.} For each image, we sample three individual latent codes from a standard normal distribution $z_k \sim \mathcal{N}(0,I_{d_k})$ and map them to intermediate latent spaces. The concatenation of all three subspaces $w=[w_1, w_2, w_3] \in \mathbb{R}^{d_1+d_2+d_3}$ serves as input to the generator $g_\eta: \mathcal{W} \rightarrow \mathcal{X}$. For \textcolor{myBlue}{GAN-inversion}, we extend our discriminator with two additional heads: one for latent representation $\hat{w}$ and another for a pixel feature vector $p$ for image encoding and pixel space reconstruction, respectively. For \textcolor{myGreen}{subspace learning}, we introduce individual mapping networks for each subspace and incorporate label information for real images through classification heads. To prevent subspace correlation, we add a penalty to the classification tasks using the average distance correlation (dCor) between the subspaces $\hat{w}_k$.}
    \label{fig:methods_gan}
\end{figure}

For the generative model, the overall goal was to generate realistic fundus images from a disentangled latent space. We considered the image generation task by learning the mapping between two domains $M: \mathcal{X} \rightarrow \mathcal{\hat{X}}$ with a latent space model. We split $M$ into two components, $M = d \circ g$, where the discriminator/encoder $d: \mathcal{X} \rightarrow \mathcal{W}$ maps from the image domain to a latent representation, and the generator $g: \mathcal{W} \rightarrow \mathcal{\hat{X}}$ maps from the latent representation to the image reconstruction.

As the backbone generative image model we selected the StyleGAN2 architecture \citep{karras2020analyzing}, a state-of-the-art generative adversarial network for high-fidelity and high-resolution images. The StyleGAN architecture has two features that makes it a good candidate for disentanglement: (1) a mapping network $m: \mathcal{Z} \rightarrow \mathcal{W}$ that maps a latent vector into an intermediate space, which then controls the styles in each convolutional layer in the generator with adaptive instance normalization, and (2) scale-specific image generation by a progressively growing generator.

We trained the generative image model with the standard minimax GAN loss together with StyleGAN's path length regularization $P_L$ \citep{karras2020analyzing} on the generator and an R1 regularizer \citep{mescheder2018training} as a gradient penalty on the discriminator for real data: 
\begin{align}
    L_{\text{GAN}}(\eta, \tau) &= \mathbb{E}_{x\in\mathbb{X}} \left[ \log(d_{\tau,0}(x)) \right]+\mathbb{E}_{z\in\mathbb{Z}} \left[ \log (1 - d_{\tau,0}(g_\eta(w))) \right] + \lambda_{P_L} P_L(\eta) - \lambda_{R_1} R_1(\tau), \\
    \eta^* &= \argmin_{\eta} \, L_\text{GAN}(\eta, \tau^*), \\
    \tau^* &= \argmin_{\tau} \, - L_\text{GAN}(\eta^*, \tau).
\end{align}
Here, $d_{\tau,0}(x)$ is the discriminator's estimate of the probability that the real data instance $x$ is real ($0 \leq d_{\tau,0}(\cdot) \leq 1$), $g_\eta(w)$ denotes the generator's output given $w$, and $d_{\tau,0}(\hat{x})$ is the discriminator's estimate of the probability that a fake instance $\hat{x} = g_\eta(w)$ is real (Fig.\,\ref{fig:methods_gan}).

As GANs were not originally designed for representation learning, they do not provide an encoder model. Therefore, we extended the StyleGAN2 architecture to embed real images in the latent space for representation learning. Inspired by the work of invGAN \citep{gosh2022inv} for GAN-inversion, we also trained the discriminator as an encoder. We added a fully connected layer to estimate the latent representation $\hat{w}$ for our first inversion loss
\begin{align}
    L_{w} (\eta, \tau) &= \mathbb{E}_{z\in\mathbb{Z}} \left[ \|w -\hat{w}\|_2^2\right] = \mathbb{E}_{z\in\mathbb{Z}} \left[  \|w - d_{\tau, 1}(g_\eta(w)) \|_2^2\right] \label{eq:gan_inv_w}.
\end{align}
We also incorporated the pixel space reconstruction loss from invGAN \citep{gosh2022inv} to further impose consistency for the GAN-inversion and to ensure that the reconstruction of an image's latent code is close to the original image in pixel space. As the definition of a meaningful distance function between real images and their reconstructions is a non-trivial task, they employ the discriminator also as a feature extractor \citep{gosh2022inv}. Therefore, we added a third head at the penultimate discriminator layer, which maps from an image to a pixel feature vector $f_p: \mathcal{X} \rightarrow p$ for our second GAN-inversion loss
\begin{align}
    L_{p} (\eta, \tau) = \mathbb{E}_{x\in\mathbb{X}} \left[ \| f_p(x) - f_p(\hat{x})\|_2^2\right] = \mathbb{E}_{x\in\mathbb{X}} \left[ \| d_{\tau, 2}(x) - d_{\tau, 2}(g_\eta(d_{\tau, 1}(x)))\|_2^2\right].\label{eq:gan_inv_p}
\end{align}
Because we have two forward passes through the decoder in Eq.\,\ref{eq:gan_inv_p}, the pixel space reconstruction loss also serves as a cycle consistency loss.

The main methodological contribution of this work is to extend the StyleGAN2 architecture (with GAN-inversion for representation learning) to an independent subspace learner. Therefore, we modified the architecture to sample individual latent codes from a standard normal distribution $z_k \sim \mathcal{N}(0,I_{d_k})$ and learned individual 8-layer mapping networks for each subspace $m_\eta^k: \mathcal{Z}_k \rightarrow \mathcal{W}_k$ (Fig.\,\ref{fig:methods_gan}, left), where each subspace can have a different dimensionality $w_k \in \mathbb{R}^{d_k}$. In setting up the generative image model, we assumed that we do not know all the underlying factors of variation or have access to their labels. Therefore, we provided the model with a content agnostic style space $w_3$ (Fig.\,\ref{fig:methods_gan}), in which the encoder/discriminator can capture further information required to reconstruct high-resolution retinal fundus images which is independent from the other content subspaces.

As in the encoder model (Sec.\,\ref{subsection:encoder model}), we encoded label information for real images by training linear classification heads $C_{\psi_k}$ for the first two subspaces (Fig.\,\ref{fig:methods_gan}, right). To avoid subspace correlation, we penalized the classification task with the mean of all distance correlation measures between subspaces $\hat{w}_k$ that the discriminator outputs.

Hence, the overall objective was 
\begin{align}
    \eta^* = &\argmin_{\eta} \, L_\text{GAN}(\eta, \tau^*) + \lambda_{w} L_{w}(\eta, \tau^*) + \lambda_{p} L_{p}(\eta, \tau^*) \label{eq:gan_generator}\\
    (\tau^*, \psi^*) = &\argmin_{\tau, \psi} \, - L_\text{GAN}(\eta^*, \tau) + \lambda_{w} L_{w}(\eta^*, \tau) + \lambda_{p} L_{p}(\eta^*, \tau) + \frac{\lambda_\text{C}}{K-1} \sum_{k=1}^{K-1} L_\text{CE}(\tau, \psi_k) + \lambda_\text{DC} L_\text{DC}(\tau) \label{eq:gan_discriminator}
\end{align}
with the generator optimization in Eq.\,\ref{eq:gan_generator} and the discriminator optimization in Eq.\,\ref{eq:gan_discriminator}. The discriminator, in its role as an encoder, also optimizes subspace losses on real images.

In practice, we optimized the objectives for the generator (Eq.\,\ref{eq:gan_generator}) and the discriminator (Eq.\,\ref{eq:gan_discriminator}) with batch estimates and stochastic gradient descent. Moreover, we used separate optimizers for generator and discriminator optimization. Appendix \ref{appendix:gan_training} offers additional information about the generative model training.

\section{Results}
We conducted empirical evaluations to assess the efficacy of our disentanglement loss in two distinct scenarios: (1) a predictive task and (2) a generative task, both performed on retinal fundus images. Our primary objective in both cases was to disentangle camera as a technical factor from the patient attributes age and ethnicity. 

To evaluate the success, we defined a quantitative measure of disentanglement. While there is no universally accepted definition of disentanglement, most agree on two main points \citep{eastwood2018a,carbonneau2022measuring}: (1) factor independence and (2) information content. Factor independence means that a factor affects only a subset of the representation space, and only this factor affects this subspace. We refer to this property as \textbf{modularity}. As another criterion for factor independence some argue that the affected representation space by a factor should be as small as possible, a principle known as \textbf{compactness}. The second agreed-upon criterion is the information content or \textbf{explicitness}. In essence, a disentangled representation should completely describe all factors of interest. To assess disentanglement, three distinct families of metrics have been proposed: intervention-based, predictor-based, and information-based metrics \citep{carbonneau2022measuring}. 

In our case, we opted for a predictor-based metric to evaluate both modularity and explicitness, excluding compactness because we fixed the subspace dimensions. For our predictor, we employed a k-nearest neighbors (kNN) classifier with k=30. Using this classifier, we generated a confusion matrix, comparing subspaces with associated labels, for example ethnicity and camera (Fig.\,\ref{fig:intro}\,\textbf{c}). Given the presence of unbalanced classes, we report the accuracy improvement over chance level accuracy. For the confusion matrix, modularity is evident by a high classification accuracy improvement for one factor (e.g. ethnicity) in only one subspace. Explicitness, on the other hand, is evident when the accuracy improvement is high across the entire matrix. Consequently, when both modularity and explicitness are effectively addressed, the confusion matrix has high values along the main diagonal as well as low values on the off-diagonals (Fig.\,\ref{fig:intro}\,\textbf{c}).

We conducted two experiments, configuring the model to learn specific subspaces. In the initial experiment, the first subspace $w_\text{age} \in \mathbb{R}^4$ encoded age information, while the second subspace $w_\text{cam} \in \mathbb{R}^{12}$ captured camera-related details. In the second experiment, we designed the first subspace $w_\text{ethn} \in \mathbb{R}^8$ to encode ethnicity information, while the second subspace focused again on camera-related features $w_\text{cam} \in \mathbb{R}^{12}$. The dimensionality of each subspace was selected based on the dimensions of the labels we aimed to encode.

We filtered for images where labels for these classes were available and ended up with 75,708 images (24,254 patients) for the first and 63,210 images (19,950 patients) for the second experiment (see Sec.\,\ref{sec:data} for details).

\subsection{Learning disentangled subspaces with an encoder model}
We first evaluated the effect of the distance correlation loss on the simplest representation learning setup for images: an encoder neural network that mapped retinal images to a latent space (Fig.\,\ref{fig:methods_encoder}). To encourage disentangled subspaces, we studied the effect of minimizing the distance correlation between them (Sec.\,\ref{subsec:disentanglement_loss}). To this end, we compared 4 runs of two model configurations on the test data: a baseline model, for which we trained only linear classifiers for subspace encoding, and a model that additionally minimized the distance correlation between subspaces. For more information about the predictive model training, see Appendix \ref{appendix:encoder_training}.

\begin{figure}[htbp]
    \centering
    \includegraphics[width=\textwidth]{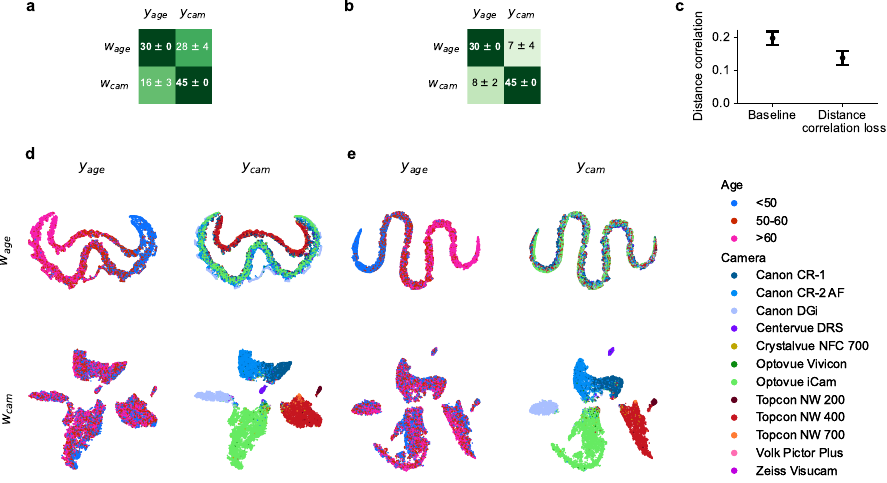}
    \caption{\textbf{Encoder disentanglement performance for age-camera setup.} We trained encoder models on retinal images, where the first subspace $w_\text{age}$ encoded age $y_\text{age}$, and the second subspace $w_\text{cam}$ encoded camera information $y_\text{cam}$. In panels \textbf{a} and \textbf{b}, we show the accuracy improvement over chance level accuracy. As a baseline model in \textbf{a}, we only trained linear classifiers on top of subspaces. In \textbf{b}, we further disentangled the subspaces using our distance correlation loss. Panel \textbf{c} shows the distance correlation measure between subspaces for both the baseline method and our proposed method for subspace disentanglement. In panels \textbf{d} and \textbf{e}, we provide t-SNE visualizations comparing a baseline model with a model with disentanglement loss, respectively.}
    \label{fig:results-encoder-disentanglement-age}
\end{figure}

Our experiments showed that the age and camera subspaces were heavily entangled in the baseline setting when disentanglement was not enforced (Fig.\,\ref{fig:results-encoder-disentanglement-age}\,\textbf{a}), such that camera information was present in the age subspace and vice versa. Entanglement was visible by relatively high classification accuracy across subspaces (Fig.\ref{fig:results-encoder-disentanglement-age}\,\textbf{a}, off-diagonals). For example, camera type could be decoded from the age subspace $w_\text{age}$ with a performance of $28\%$ above chance level, compared to $45\%$ in the camera subspace. When enforcing disentanglement by additionally minimizing the distance correlation between subspaces the cross-subspace classification performance (off-diagonals in Fig.\,\ref{fig:results-encoder-disentanglement-age}\,\textbf{b}) dropped, while the within-subspace classification was preserved (diagonals). In addition, the distance correlation between subspaces on the test data dropped as a direct consequence of our disentanglement loss (Fig.\,\ref{fig:results-encoder-disentanglement-age}\,\textbf{c}). We could also observe the disentanglement effect qualitatively in 2D t-SNE visualizations of the learned representations. In both the baseline and the disentangled model, the age subspace exhibited structure w.r.t. age, and the camera subspace encoded camera type well (diagonals in Fig.\,\ref{fig:results-encoder-disentanglement-age}\,\textbf{d, e}). In the baseline model, however, the age subspace also strongly encoded camera information (first row, second column in \textbf{d}).

\begin{figure}[htbp]
    \centering
    \includegraphics[width=\textwidth]{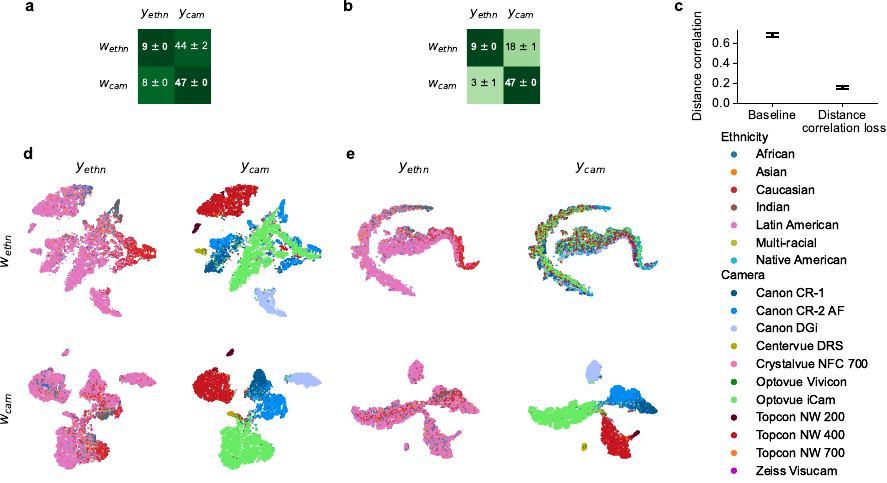}
    \caption{\textbf{Encoder disentanglement performance for ethnicity-camera setup.} We trained encoder models on retinal images, where the first subspace $w_\text{ethn}$ encoded ethnicity $y_\text{ethn}$, and the second subspace $w_\text{cam}$ encoded camera information $y_\text{cam}$. In panels \textbf{a} and \textbf{b}, we show accuracy improvement over chance level accuracy. As a baseline model in \textbf{a}, we only trained linear classifiers on top of subspaces. In \textbf{b}, we further disentangled the subspaces using our distance correlation loss. Panel \textbf{c} compares the distance correlation measure between subspaces of our two model configurations. In panels \textbf{d} and \textbf{e}, we provide t-SNE visualizations comparing a baseline model with a model with disentanglement loss, respectively.}
    \label{fig:results-encoder-disentanglement-ethn}
\end{figure}

In the second experiment, we observed an entanglement between the ethnicity and camera subspaces for the baseline method (Fig.\,\ref{fig:results-encoder-disentanglement-ethn}\,\textbf{a}). For the baseline method we observed comparable attribute performance across subspaces (columns Fig.\,\ref{fig:results-encoder-disentanglement-ethn}\,\textbf{a}). For example, camera decoding in the ethnicity subspace showed an accuracy improvement of 44\% over chance level, compared to 47\% for the dedicated camera subspace. In addition, ethnicity decoding only showed a marginal improvement over chance level, with 9\% and 8\%, respectively. This discrepancy may be due to the imbalance of our dataset, with 71\% of the patients being Latin American (Fig.\,\ref{fig:data}\,\textbf{c}). After incorporating our disentanglement loss, distance correlation dropped notably (Fig.\,\ref{fig:results-encoder-disentanglement-ethn}\,\textbf{c}). Moreover, cross-subspace classification performance decreased (off-diagonals in Fig.\,\ref{fig:results-encoder-disentanglement-ethn}\,\textbf{a}), while within-subspace classification performance remained high (diagonals in Fig.\,\ref{fig:results-encoder-disentanglement-ethn}\,\textbf{a}). The impact of the improved disentanglement was also evident in the t-SNE visualizations. The baseline model exhibited camera clustering within the ethnicity subspace (Fig.\,\ref{fig:results-encoder-disentanglement-ethn}\,\textbf{d}), a phenomenon not observed in the model with our disentanglement loss (Fig.\,\ref{fig:results-encoder-disentanglement-ethn}\,\textbf{e}).

\subsection{Ablation study for disentanglement loss} \label{subsec:results-ablation-study}
Next, we performed a sensitivity analysis for our encoder model with disentanglement loss and investigated the effects of different hyperparameters. The subspace labels used for this analysis were age and camera. In particular, we examined the subspace dimensionality, the batch size, and $\lambda_{DC}$ (Eq.\,\ref{eq:encoder_loss}).

The initial hyperparameters were set as follows: a 4-dimensional age subspace $w_\text{age}$, a 12-dimensional camera subspace $w_\text{cam}$, a batch size of 512, and $\lambda_{DC}=0.5$. Subsequently, each hyperparameter was individually varied, and for each configuration, we trained four models with different random initializations. All evaluation metrics were then reported on the test set.

\begin{figure}[htbp]
    \centering
    \includegraphics[width=\textwidth]{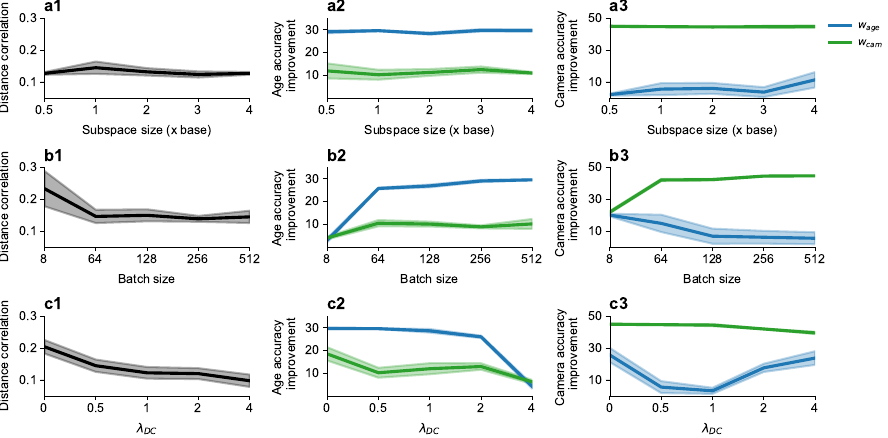}
    \caption{\textbf{Impact of hyperparameters on disentanglement performance.} In \textbf{a1-a3}, we present performance variations across different subspace sizes. Specifically, \textbf{a1} illustrates the distance correlation between subspaces, while \textbf{a2} and \textbf{a3} depict the accuracy improvement over chance level for age and camera, respectively. In \textbf{b1-b3}, we examine disentanglement performance under varying batch sizes. \textbf{c1-c3} focus on the impact of different distance correlation loss weights on disentanglement performance. We evaluated all metrics on the test data over four models with different initializations.}
    \label{fig:results-encoder-ablation}
\end{figure}

The robustness of our disentanglement loss was evident across varied subspace sizes (Fig.\,\ref{fig:results-encoder-ablation}\,\textbf{a1-a3}). Interestingly, there were no clear trends when we altered the subspace size by multiplying our initial sizes with different factors. Notably, one exception was observed in the accuracy improvement for the camera within the age subspace, particularly when employing a multiplier of 4 (Fig.\,\ref{fig:results-encoder-ablation}\,\textbf{a3}, blue curve). In this case, the disentanglement problem becomes more challenging as we need to estimate the distance correlation between a 16- and a 48-dimensional subspace with a limited batch size of 512. 

Smaller batch sizes impacted the disentanglement performance (Fig.\,\ref{fig:results-encoder-ablation}\,\textbf{b1-b3}). This effect was notable with the higher distance correlation values and worse decoding quality for a batch size of 8 compared to larger batch sizes (Fig.\,\ref{fig:results-encoder-ablation}\,\textbf{b1-b3}). In particular, for the age subspace, a batch size of 8 and 64 proved to be insufficient, as shown by the camera decoding not reaching a possible minimum observed with larger batch sizes (Fig.\,\ref{fig:results-encoder-ablation}\,\textbf{b3}, blue curve). 
 
In our last ablation experiment, we explored the loss weighting term $\lambda_{DC}$ for distance correlation minimization (Fig.\,\ref{fig:results-encoder-ablation}\,\textbf{c1-c3}). As expected, the distance correlation measure between subspaces dropped with larger $\lambda_{DC}$ (Fig.\,\ref{fig:results-encoder-ablation}\,\textbf{c1}). At the same time, the decoding accuracy for age from the age subspace (Fig.\,\ref{fig:results-encoder-ablation}\,\textbf{c2}, blue curve) collapsed for $\lambda_{DC} \geq 2$, leading to a drop in decoding performance below 10\%. In contrast, camera decoding (Fig.\,\ref{fig:results-encoder-ablation}\,\textbf{c2}, green curve) from the age subspace becomes less accurate as soon as the disentanglement loss is introduced, indicating that the proposed method does what it is supposed to. For the camera subspace in (Fig.\,\ref{fig:results-encoder-ablation}\,\textbf{c3}), the camera decoding accuracy remains high (green), and the age decoding accuracy initially drops (blue), in line with the goals of the distance correlation loss. However, the ablation study showed that the hyperparameter optimization of $\lambda_{DC}$ is essential, as over-weighting can lead to a complete collapse of the subspace encoding (Fig.\,\ref{fig:results-encoder-ablation}\,\textbf{c2,c3}).

\subsection{Learning disentangled subspaces with a generative model} \label{subsec:results-disentanglement-gan}
In the second part of the experiments, we extended our predictive model to a generative model, aiming to generate realistic retinal images from a disentangled latent space (Sec.\,\ref{subsection:generative model}). 

As with the predictive model, we encoded label information into latent subspaces by optimizing linear classifiers on distinct tasks. However, in the generative framework, we also introduced an additional latent subspace dedicated to encoding image style,  $w_\text{style} \in \mathbb{R}^{16}$ (Fig.\,\ref{fig:methods_gan}). To ensure independence between all subspaces, we incorporated distance correlation minimization. Consequently, we aimed to minimize the distance correlation between three different subspace combinations. Notably, while we did not have an additional classification head for the style space, we specifically enforced independence from the other subspaces, ensuring that it remained uncorrelated with age, ethnicity, or camera information.

We ran 5 models of each configuration -- a baseline model and a model with distance correlation loss. However, during our experiments, we had to discard one model for evaluation as it did not converge properly, leaving us with 4 models for evaluation on the test data.

\begin{figure}[t]
    \centering
    \includegraphics[width=\textwidth]{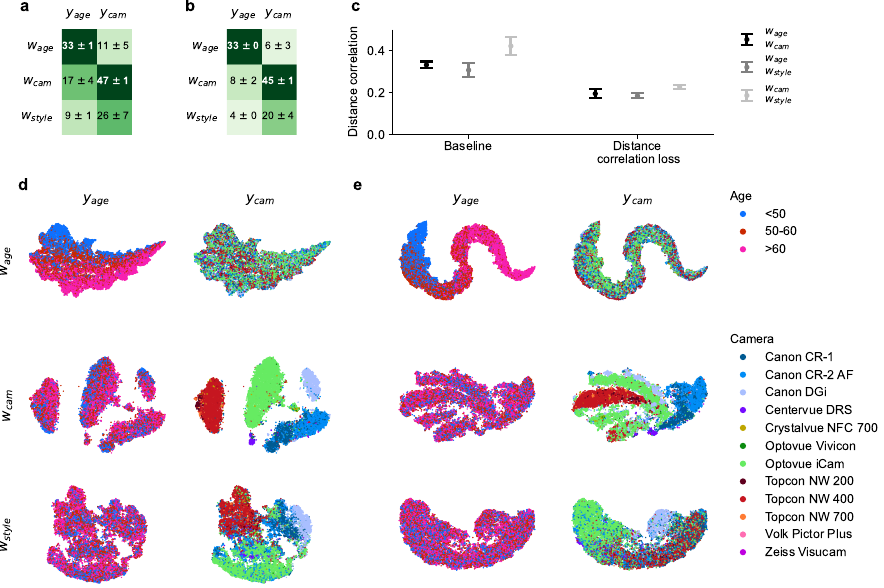}
    \caption{\textbf{Disentanglement performance for age-camera setup.} We trained generative models on retinal images, featuring three subspaces: $w_\text{age}$ encoding the age class $y_\text{age}$, $w_\text{cam}$ encoding the camera class $y_\text{cam}$, and $w_\text{style}$ serving as a style subspace. In \textbf{a} and \textbf{b}, we report kNN classifier accuracy improvement over chance level accuracy. The baseline model in \textbf{a} involved training linear classifiers on the first two subspaces. In \textbf{b}, we additionally disentangled the subspaces using our distance correlation loss. Panel \textbf{c} compares the distance correlation measure between subspaces of our two model configurations. Panels \textbf{d} and \textbf{e} present t-SNE visualizations of one baseline model and one trained generative model with the disentanglement loss, respectively.}
    \label{fig:results-gan-disentanglement-age}
\end{figure}

Our disentanglement loss also showed an effect when applied to a generative image model (Fig.\,\ref{fig:results-gan-disentanglement-age}). While the baseline model for the generative model already demonstrated relatively good disentanglement performance, the introduction of the distance correlation loss further enhanced the separation of subspaces.
For example, for the baseline model, age information could be decoded from the camera subspace and vice versa (Fig.\,\ref{fig:results-gan-disentanglement-age}\,\textbf{a}, off-diagonals). Additionally, age, and particularly camera information, were still contained within the style subspace of the baseline model (last row, Fig.\,\ref{fig:results-gan-disentanglement-age}\,\textbf{a}). Upon incorporating the distance correlation loss, cross-subspace classification performance (off-diagonals and last row, Fig.\,\ref{fig:results-gan-disentanglement-age}\,\textbf{b}) dropped, while within-subspace classification was preserved (diagonals). Consequently, the distance correlation between subspaces on the test data decreased (Fig.\,\ref{fig:results-gan-disentanglement-age}\,\textbf{c}), highlighting the effect of our disentanglement loss. The qualitative impact of disentanglement was further evident in t-SNE visualizations: for example, the camera labels clustered in the style space for the baseline model, but the clustering became more mixed up when we incorporated the distance correlation loss (Fig.\,\ref{fig:results-gan-disentanglement-age}\,\textbf{d,e}). 

\begin{figure}[t]
    \centering
    \includegraphics[width=\textwidth]{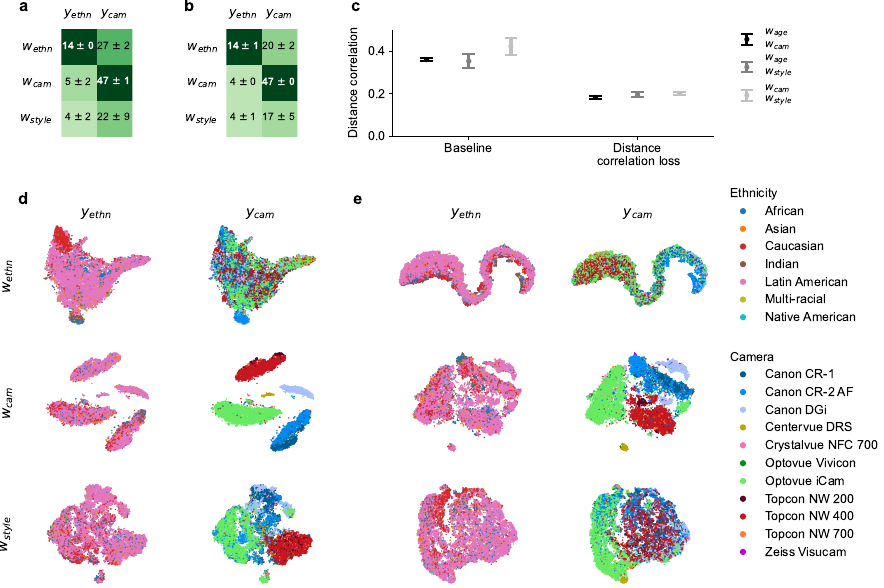}
    \caption{\textbf{Disentanglement performance for ethnicity-camera setup.} We trained generative models on retinal images, featuring three subspaces: $w_\text{ethn}$ encoding the ethnicity class $y_\text{ethn}$, $w_\text{cam}$ encoding the camera class $y_\text{cam}$, and $w_\text{style}$ serving as the style subspace. In \textbf{a} and \textbf{b}, we report kNN classifier accuracy improvement over chance level accuracy. The baseline model in \textbf{a} involved training linear classifiers on the first two subspaces. In \textbf{b}, we additionally disentangled the subspaces using our distance correlation loss. Panel \textbf{c} compares the distance correlation measure between subspaces of our two model configurations. Panels \textbf{d} and \textbf{e} present t-SNE visualizations of one baseline model and one trained generative model with the disentanglement loss, respectively.}
    \label{fig:results-gan-disentanglement-ethn}
\end{figure}

In our second experiment, our objective was to disentangle ethnicity, camera, and style within three subspaces of a generative model (Fig.\,\ref{fig:results-gan-disentanglement-ethn}). Again the baseline model for the generative model already demonstrated relatively good disentanglement performance compared to the encoder model (Fig.\,\ref{fig:results-encoder-disentanglement-ethn}). However, the application of our disentanglement loss still demonstrated several effects. For instance, the accuracy improvement for camera decoding decreased by 7\% for the ethnicity subspace and by 5\% for the style subspace on average (Fig.\,\ref{fig:results-gan-disentanglement-ethn}, \textbf{a} versus \textbf{b}). As a direct consequence of minimizing distance correlation between subspaces, this measure consistently decreased for all subspace combinations (Fig.\,\ref{fig:results-gan-disentanglement-ethn}\,\textbf{c}). The t-SNE visualizations not only qualitatively showed the classification and disentanglement performance (Fig.\,\ref{fig:results-gan-disentanglement-ethn}\,\textbf{d,e}), but also pointed us to a correlation in the ethnicity subspace between Indian patients (first row, first column) and the Canon CR-2 AF camera (first row, second column), as they both co-occur in the same clusters, even after incorporating our disentanglement loss. Another interesting finding was that the ethnicity decoding performance from the ethnicity subspace exhibited a 5\% improvement for the generative model compared to the encoder model (Fig.\,\ref{fig:results-encoder-disentanglement-ethn}\,\textbf{a,b}). This disparity may suggest that the generative model serves as a better feature extractor because of a better learning bias or a greater model capacity.

\subsection{Realistic image generation and reconstruction}
Next, we evaluated the image quality and the reconstruction performance of our generative model. As a baseline we chose a standard StyleGAN2 model with GAN-inversion losses, which we compared to models incorporating our additional losses for independent subspace learning. The baseline model was trained with a 32-dimensional latent space and a single mapping network. For the comparison models, we chose our $\mathcal{W}$-spaces to be either 32- or 36-dimensional for the age-camera-style and ethnicity-camera-style setups, respectively. All models generated retinal fundus images with a resolution of $256\times 256$ pixels.

\begin{figure}[htbp]
    \centering
    \includegraphics[width=\textwidth]{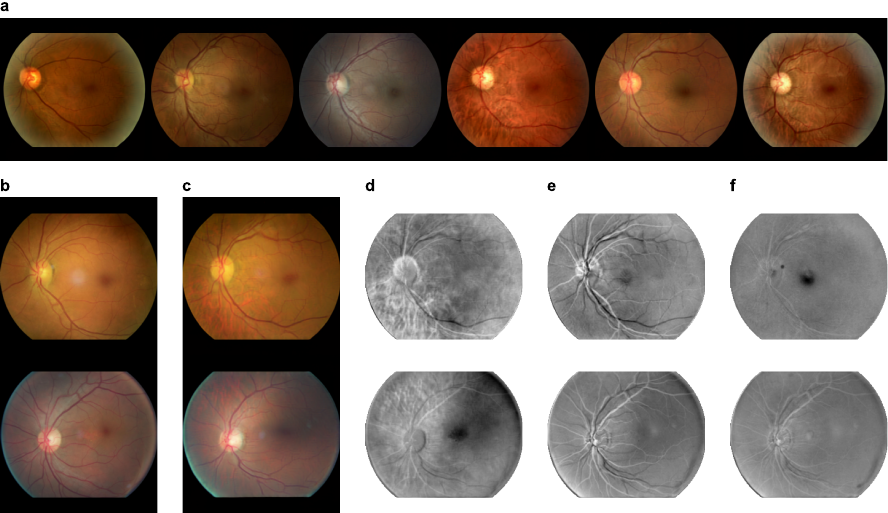}
    \caption{\textbf{Qualitative image generation and inversion performance.} We present a qualitative assessment of image generation performance for one model of configuration (C) (Tab.\,\ref{tab:results_image_quality}). In panel \textbf{a}, we show images generated from randomly sampled latent vectors. Panel \textbf{b} features retinal image samples from the test dataset. In panel \textbf{c}, we display the corresponding reconstructions produced by the generative model. Additionally, we show the difference maps for each of the RGB channels in \textbf{d-f}.}
    \label{fig:results-gan-image-quality}
\end{figure} 

We were able to generate realistic images and accurately reconstruct real images: generated retinal fundus images from randomly sampled latent vectors showed retinas of varying pigmentation with an optic disc and macula (Fig.\,\ref{fig:results-gan-image-quality}\,\textbf{a}). The fine vascular structures showed different patterns, resulting in a general appearance similar to the original data distribution. However, on closer inspection, the generated vasculature was not always intact. Furthermore, the inversion performance of the model was qualitatively robust, especially for coarse structures. A visual comparison of some retinal images (Fig.\,\ref{fig:results-gan-image-quality}\,\textbf{b}) with their reconstructions (Fig.\,\ref{fig:results-gan-image-quality}\,\textbf{c}) showed similarities in pigmentation, shape, position of the macula, and general vascular structure. However, detailed examination of the difference maps (Fig.\,\ref{fig:results-gan-image-quality}\,\textbf{d-f}) revealed some inaccuracies in the precise reconstruction of the vasculature, especially for thin vessels.

In a quantitative evaluation we confirmed that the image quality was not affected by additional constraints for learning disentangled subspaces (Tab.\,\ref{tab:results_image_quality}). To measure image quality, we used Frechet Inception Distance (FID) \citep{heusel2018, parmar2021cleanfid}, which assesses differences between the distribution densities of real and fake images in the feature space of an InceptionV3 classifier \citep{simonyan2015}. For the baseline model, we measured an FID score of 14 (Tab.\,\ref{tab:results_image_quality}, model A). Subsequent models incorporating subspace classifiers showed FID scores similar to the baseline, with no notable decrease for models also incorporating the distance correlation loss (Tab.\,\ref{tab:results_image_quality}, B versus C or D versus E). 

Furthermore, we examined how additional subspace losses affected the image reconstruction performance, by reporting our inversion losses $L_w$ and $L_p$ for image encoding and pixel space reconstruction, respectively. As expected, the image encoding loss $L_w$ consistently showed a slight degradation for models that we jointly optimized with our disentanglement loss (Tab.\,\ref{tab:results_image_quality}, model C and E). In contrast, the pixel space reconstruction loss $L_p$ remained comparable across all configurations. However, $L_p$ showed worse mean performance with high variances for some model configurations (Tab.\,\ref{tab:results_image_quality}, model A and D), suggesting that this loss could potentially benefit from a higher weight, which was also visible in the qualitative reconstruction difference maps in pixel space (Fig.\,\ref{fig:results-gan-image-quality}\,\textbf{d-f}).

\begin{table*}[h]
    \centering
     \caption{\textbf{Image quality and inversion performance.} For different model configurations we report the FID scores between the training dataset and the generated images, along with the inversion losses $L_w$ and $L_p$. All metrics are reported on the test set as the average over four models with different weight initializations.}
    \begin{tabular}{l | l l l} 
        \hline
        Model configuration & FID $\downarrow$ & $L_{w}$ $\downarrow$ & $L_{p}$ $\downarrow$\\
        \hline
        (A)  Baseline GAN with inversion losses & 14 $\pm$ 3 &  0.025 $\pm$ 0.01 & 0.010 $\pm$ 0.010 \\
        \hline
        (B)  Model A + subspace classifiers (age, camera) & 13 $\pm$ 3 &  0.028 $\pm$ 0.01 & 0.002 $\pm$ 0.002 \\
        (C)  Model B + distance correlation loss & 11 $\pm$ 2 &  0.031 $\pm$ 0.01 & 0.006 $\pm$ 0.006 \\
        (D)  Model A + subspace classifiers (ethnicity, camera) & 12 $\pm$ 2 &  0.025 $\pm$ 0.01 & 0.010 $\pm$ 0.020 \\
        (E)  Model D + distance correlation loss & 13 $\pm$ 2 &  0.027 $\pm$ 0.01 & 0.002 $\pm$ 0.001 \\
        \hline
    \end{tabular}
    \label{tab:results_image_quality}
\end{table*}

\subsection{Controllable fundus image generation}
In our final experiment, we explored how disentangled subspaces of a generative model influence the generation of retinal fundus images. We started by embedding the original images into latent spaces using our generative model. Because we trained our models with a disentanglement loss, we were able to split the latent spaces of our models into three independent subspaces: the first subspace represented a patient attribute (age or ethnicity), the second subspace encoded the camera, and the third subspace represented style information. Subsequently, we performed subspace swapping by exchanging the subspace embeddings of different retinal images, allowing us to observe the effect of each subspace on image generation.

\begin{figure}[htbp]
    \centering
    \includegraphics[width=\textwidth]{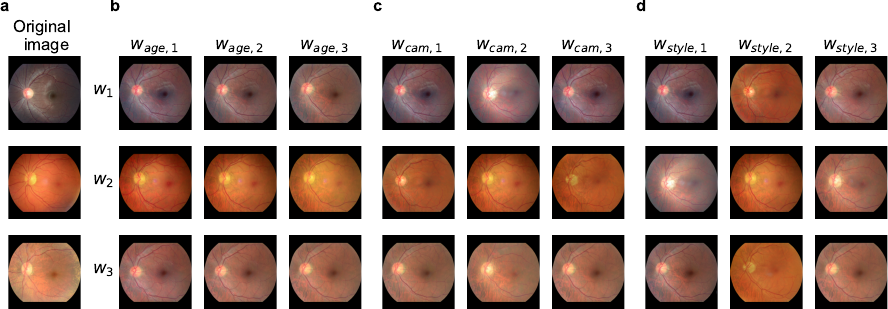}
    \caption{\textbf{Reconstructions of swapped latent subspaces for age-camera-style disentanglement.} In panel \textbf{a}, original fundus images are presented alongside their respective reconstructions in panels \textbf{b-d}. Each column in panels \textbf{b-d} corresponds to the exchange of age, camera, or style subspaces between the image encodings of panel \textbf{a}. The main diagonals in panels \textbf{b-d} show reconstructions of the images in \textbf{a}, while the off-diagonals illustrate reconstructions where one subspace has been replaced with the subspace indicated in the column header.}
    \label{fig:results-gan-swap-subspaces-age}
\end{figure}

\begin{figure}[htbp]
    \centering
    \includegraphics[width=\textwidth]{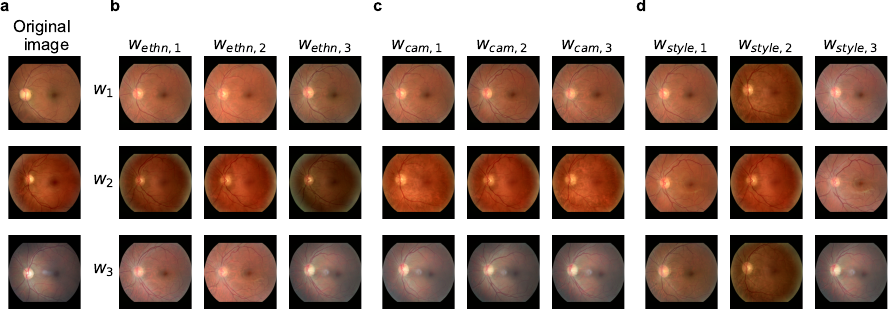}
    \caption{\textbf{Reconstructions of swapped latent subspaces for ethnicity-camera-style disentanglement.} In panel \textbf{a}, original fundus images are presented alongside their respective reconstructions in panels \textbf{b-d}. Each column in panels \textbf{b-d} corresponds to the exchange of ethnicity, camera, or style subspaces between the image encodings of panel \textbf{a}. The main diagonals in panels \textbf{b-d} show reconstructions of the images in \textbf{a}, while the off-diagonals illustrate reconstructions where one subspace has been replaced with the subspace indicated in the column header.}
    \label{fig:results-gan-swap-subspaces-ethnicity}
\end{figure}

The subspace swapping yielded matrices of image reconstructions from different latent subspace combinations (Fig.\,\ref{fig:results-gan-swap-subspaces-age} and Fig.\,\ref{fig:results-gan-swap-subspaces-ethnicity}). In each column, one fundus image subspace was replaced with another one. Consequently, the main diagonals in the resulting matrices represent reconstructions of the original images, while the off-diagonals show reconstructions where one subspace was replaced by the subspace of another retinal image.

For the qualitative analysis, we selected the best of our age-camera-style models based on disentanglement performance. The original images were randomly sampled from the test set, with the following age and camera labels (Fig.\,\ref{fig:results-gan-swap-subspaces-age}, \textbf{a}): (1) 22, Topcon NW 400, (2) 52, Optovue iCam, (3) 76, Canon CR-1. By swapping different subspaces, we generated matrices of image reconstructions from different latent subspace combinations (Fig.\,\ref{fig:results-gan-swap-subspaces-age}\,\textbf{b-d}).
Swapping the age subspace between image encodings for reconstruction led to subtle feature changes (Fig.\,\ref{fig:results-gan-swap-subspaces-age}\,\textbf{b}). Notably, bright and reflective features around the thickest vasculature were observed, which tended to fade for age subspaces belonging to an older person (Fig.\,\ref{fig:results-gan-swap-subspaces-age}\,\textbf{b}, first row). These bright and sheen structures are particularly common in the young population and tend to decrease with age \citep{williams1982}.
We also investigated the impact of changes in the camera subspace on fundus image generation (Fig.\,\ref{fig:results-gan-swap-subspaces-age}\,\textbf{c}). In this model, each column representing a different camera subspace displayed subtle feature changes, occasionally affecting identity features such as the optic disc and vasculature (second column, Fig.\,\ref{fig:results-gan-swap-subspaces-age}\,\textbf{c}). Finally, we swapped the style subspaces, which were the largest subspaces with 16 dimensions each (Fig.\,\ref{fig:results-gan-swap-subspaces-age}\,\textbf{d}). The idea was to store all remaining information relevant to the image reconstruction in this subspace. As intended, we observed a similar appearance of vasculature, optic disc, and fundus pigmentation in each style column (Fig.\,\ref{fig:results-gan-swap-subspaces-age}\,\textbf{d}, columns).

\begin{table}[h]
    \centering
     \caption{\textbf{Quantitative age space performance.} In an image classification setting, we demonstrate the influence of the age subspace on image generation by capturing relevant features. We trained image classifiers using reconstructions from our 4 GAN models on the 3-class age problem. Subsequently, we tested the classifiers on the 3-class problem ($<50$, $50-60$, $\geq60$) and a simpler 2-class problem ($\leq 40$, $\geq 65$). Here, we present the mean and standard deviation of classification performances, illustrating how age predictions correctly changed when we swapped age subspaces.}
    \begin{tabular}{c  l | c} 
        \hline
        classification task & test configuration &  accuracy in \% \\
        \hline
        \multirow{3}{*}{3-class problem} & standard & 70 $\pm$ 1\\
        & swapped age subspaces and labels & 66 $\pm$ 2 \\
        & swapped age subspaces, original labels & 40 $\pm$ 0\\
        \hline 
        \multirow{3}{*}{2-class problem} & standard & 87 $\pm$ 4\\
        & swapped age subspaces and labels & 85 $\pm$ 6\\
        & swapped age subspaces, original labels & 47 $\pm$  2\\
    \end{tabular}
    \label{tab:results_quantitative_age_space_swaps}
\end{table}

We also provide quantitative evidence that the age subspace indeed influences image generation with relevant age features. To quantitatively assess the controllable image generation, we trained an image classifier on the task of predicting age from image reconstructions of our four GAN models. For detailed information about the training and testing procedures, please refer to Appendix \ref{appendix:encoder_age_space}. We tested the classifier models on a 3-class problem ($<50$, $50-60$, $\geq60$) and a simpler 2-class problem distinguishing young from old patients ($\leq 40$, $\geq 65$). These two classification problems are almost balanced, with chance level accuracies of 37\% for the 3-class problem and 57\% for the 2-class problem.
For testing, we evaluated the classification model in three scenarios: standard evaluation on test data, evaluation on reconstructed images from swapped age subspaces and their corresponding new age labels, and as a reference, we reconstructed images from swapped age subspaces but evaluated the classification performance against the original age labels. The third evaluation method ensured that the age information was not contained in the other subspaces.
Comparing the three scenarios for both classification problems, we demonstrated that the age subspace indeed influenced image generation with the relevant features, as age predictions changed correctly when we swapped age subspaces. For example, in the 3-class problem, the performance dropped only from 70\% to 66\% for swapped age subspaces, compared to 40\% for the wrong age label assignments (Tab.\,\ref{tab:results_quantitative_age_space_swaps}). Similarly, in the 2-class problem, we observed a drop from only 87\% to 85\%, compared to an accuracy of 47\% for the wrong labels (Tab.\,\ref{tab:results_quantitative_age_space_swaps}).

In a second experiment, we explored the performance of subspace swaps using our best ethnicity-camera-style model (Fig.\,\ref{fig:results-gan-swap-subspaces-ethnicity}). For this experiment, we randomly sampled three patients from the test set with the following ethnicity and camera labels (Fig.\,\ref{fig:results-gan-swap-subspaces-ethnicity}\,\textbf{a}): (1) African, Topcon NW 400, (2) Caucasian, Optovue iCam, (3) Latin American, Canon CR-1.
Swapping ethnicity subspaces had a qualitatively stronger visual effect on the image generation than swapping age subspaces (Fig.\,\ref{fig:results-gan-swap-subspaces-ethnicity}\,\textbf{b}). In this case, the fundus pigmentation changed for different ethnicity subspaces, which is biologically plausible given the association between retinal pigmentation and ethnicity \citep{Rajesh2023}. Interestingly, some ethnicity subspace swaps also altered other patient features, such as the appearance of the vasculature (Fig.\,\ref{fig:results-gan-swap-subspaces-ethnicity}\,\textbf{b}, last column). The camera subspace swaps exhibited only subtle visual feature changes, occasionally affecting patient features like the vasculature or the optic disc (Fig.\,\ref{fig:results-gan-swap-subspaces-ethnicity}\,\textbf{c}, first column). When we changed the style subspace, other patient features (vasculature, optic disc) changed as expected (Fig.\,\ref{fig:results-gan-swap-subspaces-ethnicity}\,\textbf{d}, rows).

\section{Discussion}
In this work, we focused on bridging the gap between existing generative models and disentangled representation learning for high-quality fundus image generation. We addressed the challenge of solving three joint tasks: (1) disentangling representations, (2) preventing shortcut learning, and (3) enabling controllable high-resolution image generation. To achieve this, we drew on recent advances in generative modeling and introduced a subspace GAN as a controllable population model for retinal fundus images. Our approach included the use of a disentanglement loss based on distance correlation. Through both qualitative and quantitative analyses, we successfully demonstrated the effectiveness of our disentangled subspaces within a generative model. Our approach is not only applicable to the field of medical imaging. The general prerequisites for our method can also be fulfilled in other areas: prior knowledge of confounding factors in the image generation process and labels for the primary task and the confounding factor.

While our model shows excellent image generation performance with disentangled latent representations, some limitations should be acknowledged. One limitation is that the image reconstructions of our generative model showed deficits in capturing fine details, which is particularly evident in the accuracy of the vessel reconstructions. Interestingly, we also found that inherent correlations within the dataset pose a challenge for disentanglement. Our experiments revealed that disentangling age from the camera was more straightforward than disentangling ethnicity from the camera, and that the disentanglement performance seemed to be limited by the strong correlation between ethnicity and camera in the EyePACS dataset \citep{traeuble2021disentangled, funke2022disentanglement, roth2023disentanglement}. In addition, it was challenging to disentangle the style space from the camera information. This difficulty may stem from the fact that these represent the largest subspace combinations, spanning 12- and 16-dimensional spaces, thus posing the greatest challenge for distance correlation estimation. Additionally, it is conceivable that camera features are correlated with other dataset features relevant to retinal fundus image generation, further complicating disentanglement. In general, approaches such as ours depend on the dataset distribution at hand and are therefore limited by inherent correlations. For many retrospective datasets obtained from routine care, patient demographics are determined by the visitors to the hospital. With our approach, we offer an algorithmic solution to disentangle (correlated) factors in the dataset to a certain degree.

Moreover, our distance correlation-based disentanglement loss introduced some challenges. First, our model architecture made it necessary to find a suitable balance between the disentanglement loss, the feature encoding, and the image generation task, and over-weighting the disentanglement loss led to poor encoding and image generation performance. Second, our disentanglement loss also introduced some technical challenges. Notably, distance correlation computation is sensitive to the batch size as a hyperparameter, necessitating the introduction of a ring buffer for multi-GPU training (Appendix \ref{appendix:gan_training}). However, the size of the ring buffer was constrained by the need to balance stored batches from the past and a model that is updated every step. Furthermore, driven by distance correlation estimation, we used small latent space sizes that still enable high-quality retinal image generation. Consequently, distance correlation poses a limitation on scaling up the latent space. Larger latent space sizes would require correspondingly larger batch sizes for accurate distance correlation estimation. This interdependence prompts a deeper exploration of the dynamics between the data ring buffer, the latent space size, and the number of latent subspaces to improve our understanding of their joint impact on model performance.

To some extent, our method is scalable. Our method can be trained on larger datasets because the model training time scales linearly with the size of the dataset. In addition, our method can handle images of different resolutions well, as we show in the Appendix with Fig.\,\ref{fig:appendix-image-resolution}. Furthermore, our method can also handle combinations of multiple attributes, as we show in Fig.\,\ref{fig:appendix-multiple-subspaces}. Of course, the combinatorial complexity will eventually prohibit more attributes.

In this study, our focus was exclusively on healthy fundus images due to the additional complexity that would have been introduced by the reconstruction of disease features. However, the issue of spurious correlation between disease and technical features is a practical challenge that we aim to tackle in future work. An additional latent disease subspace, e.g. learned with the EyePACS diabetic retinopathy (DR) labels, could be an interesting clinical test scenario for our work. Here, we could investigate whether a disentangled DR subspace is more robust for disease prediction on a shifted test distribution, e.g. from another hospital. However, in this study we only worked with healthy fundus images due to the imbalanced dataset. In another study \citep{ilanchezian2023counterfactualsretinal}, we used diffusion models for counterfactual image generation (without learning any representations), and disease features could be generated here with massive oversampling of the diseased classes, so this may be a way forward here as well. 

Moreover, alternative measures to distance correlation exist, such as maximum mean discrepancy (MMD) \citep{sejdinovic2013eq}, MI bounds, or the use of adversarial classifiers, which can offer advantages, especially in terms of batch size or convergence time. We intend to investigate and compare these methods in the context of subspace disentanglement for medical images in future studies. 

In this work, we relied on labeled data to encode information into subspaces. Therefore, another compelling avenue for future research is to explore weakly supervised learning approaches, where we could apply a classification loss only on a subset of the training data, in addition to our disentanglement loss. This exploration could potentially yield effective results even on minimally labeled data. 

Lastly, the main contribution of this work is to extend an inverted GAN in general as an independent subspace learner that is applicable beyond the domain of fundus images. However, since fundus images have very specific properties (circular shape, fine vascular tree structure, optic disc, macula, etc.), biologically plausible biases could improve fundus image generation, which we leave for future work.

In conclusion, our study presents an interpretable solution for modeling simple causal relationships for a medical dataset, aiming to mitigate shortcut learning arising from spurious correlations. We hope that this research provides a foundational approach for tackling confounded datasets and contributes to the ongoing exploration of methodologies towards comprehending the interplay between patient attributes and technical confounders.

\section*{Acknowledgements}
We thank Francesco Locatello for discussing details about the background on causal representation learning and we thank Holger Heidrich for discussing project directions and implementation details. We also thank Patrick Köhler, Kyra Kadhim, Simon Holdenried-Krafft, and Jeremiah Fadugba for helpful discussions and Kyra Kadhim, Julius Gervelmeyer, and Holger Heidrich for comments on the manuscript. This project was supported by the Hertie Foundation and by the Deutsche Forschungsgemeinschaft under Germany’s Excellence Strategy with the Excellence Cluster 2064 ``Machine Learning — New Perspectives for Science'', project number 390727645. This research utilized compute resources at the Tübingen Machine Learning Cloud, INST 37/1057-1 FUGG. PB is a member of the Else Kr\"oner Medical Scientist Kolleg ``ClinbrAIn: Artificial Intelligence for Clinical Brain Research''. LK is supported by the Diabetes Center Berne. The authors thank the International Max Planck Research School for Intelligent Systems (IMPRS-IS) for supporting SM.

\bibliographystyle{plainnat}
\bibliography{references.bib}

\newpage
\begin{appendices}

\renewcommand\thefigure{\thesection.\arabic{figure}} 
\setcounter{figure}{0}   

\section{Toy example for disentanglement loss}\label{sec:appendix_toy_example}
\begin{figure}[htbp]
    \centering
    \includegraphics[width=\textwidth]{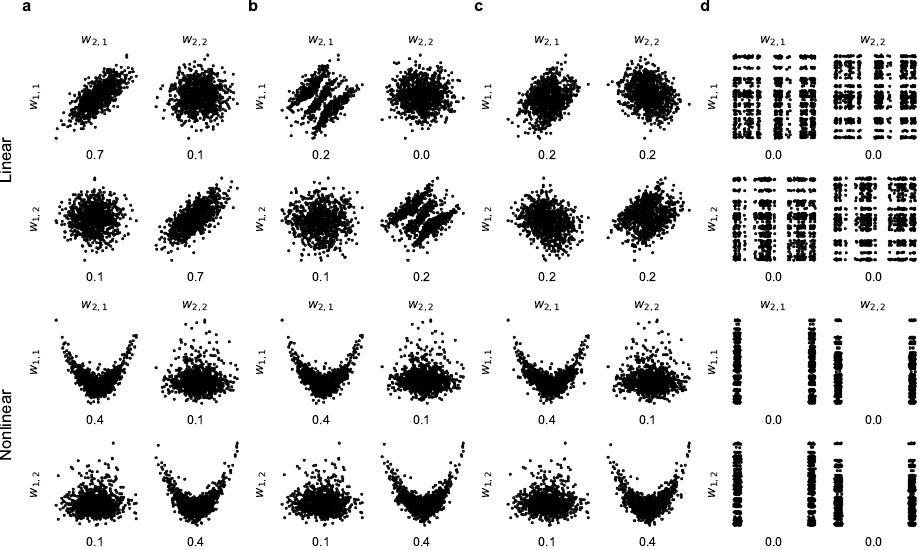}
    \caption{\textbf{Toy example for different dependence measures.} We sampled 1,000 points from two 2-dimensional subspaces $w_1 = [ w_{1,1}, w_{1,2} ] \in \mathbb{R}^2$ and $w_2 = [ w_{2,1}, w_{2,2} ] \in \mathbb{R}^2$ and created dependencies between the main diagonals in the pair plots of \textbf{a}. We defined an optimization problem where we changed point coordinates to minimize various dependence measures. Below the points we show distance correlation as a nonlinear dependence measure. In panels \textbf{b} and \textbf{c} we show the displacement of points when minimizing the linear dependence measures GMI and C-GMI, respectively. In panel \textbf{d} we see that, compared to GMI and C-GMI, distance correlation can also effectively disentangle nonlinear dependencies between subspaces.}
    \label{fig:appendix-toy-example-moving-points}
\end{figure}

Using a simple toy example, we showed that distance correlation can measure nonlinear dependencies between variables compared to two linear dependence measures. We sampled 1,000 points from two 2-dimensional subspaces $w_1 = [ w_{1,1}, w_{1,2} ] \in \mathbb{R}^2$ and $w_2 = [ w_{2,1}, w_{2,2} ] \in \mathbb{R}^2$ and created dependencies between them (Fig.\,\ref{fig:appendix-toy-example-moving-points}\,\textbf{a}, main diagonals in the pair plots). We defined an optimization problem, where points should be moved in order to minimize different dependence measures (Fig.\,\ref{fig:appendix-toy-example-moving-points}\,\textbf{b-d}). We optimized point coordinates with PyTorch \citep{paszke2019pytorch} and an Adam optimizer ($\beta_1 = 0.9, \beta_2 = 0.99,\epsilon=10^{-8}$) with a learning rate of 0.05 for 500 steps. When we optimized for the linear dependence measures Gaussian mutual information (GMI) and collapsed Gaussian mutual information (C-GMI), distance correlation as a nonlinear dependence measure only decreased for linear dependencies (Fig.\,\ref{fig:appendix-toy-example-moving-points}\,\textbf{b-c}, distance correlation values below points). GMI and C-GMI are two mutual information estimators that assume that the subspace vectors $w_1$ and $w_2$ follow a multivariate normal distribution (see \ref{appendix:appendix_gmi} for details). However, when we optimized for distance correlation, the dependence measure also dropped for nonlinear dependencies (Fig.\,\ref{fig:appendix-toy-example-moving-points}\,\textbf{d}, distance correlation values below points).

\subsection{Gaussian mutual information} \label{appendix:appendix_gmi}
Mutual information (MI) between two random vectors is defined as 
\begin{align*}
    \text{MI}(w_1,w_2) = D_\text{KL} \left( p(w_1,w_2) || p(w_1) p(w_2) \right).
\end{align*}
When we assume the subspace vectors follow a multi-variate normal distribution $w_1 \sim \mathcal{N}(\mu_1, \Sigma_1)$ and $w_2 \sim \mathcal{N}(\mu_2, \Sigma_2)$, the MI simplifies to 
\begin{align*}
    \text{GMI}(w_1,w_2) &= D_\text{KL} \left( p(w_1,w_2) || p(w_1) p(w_2)\right)  \\
    \text{GMI}(w_1,w_2) &= \frac{1}{2} \left(\log\left( \frac{\det\Sigma_{w_1} \det\Sigma_{w_2}}{\det\Sigma}\right) \right)
\end{align*}
which we call a Gaussian mutual information (GMI).
Here, $\Sigma$ is the covariance matrix of the joint distribution $p(w_1,w_2)$
\begin{align*}
    \Sigma &= \begin{pmatrix}
        \Sigma_{w_1} & \Sigma_{w_1,w_2} \\
        \Sigma_{w_1,w_2} & \Sigma_{w_2}
    \end{pmatrix}
\end{align*}

Approximating this measure over batches leads to numerically instabilities for covariance matrix estimation. Hence, we simplified the computation by taking the sum of the variables over their feature dimension $\hat{w_1} = \Sigma_{i=0}^{d_1} w_{1,i}$ and $\hat{w_2} = \Sigma_{i=0}^{d_2} w_{2,i}$. Since we collapse feature dimensions here, we name this measure collapsed Gaussian mutual information (C-GMI).
\begin{align*}
    \text{C-GMI}(w_1,w_2) = \frac{1}{2} \log\left( \frac{\text{var}(\hat{w_1}) \text{var}(\hat{w_2})} {\text{var}(\hat{w_1}) \text{var}(\hat{w_2}) - \text{cov}(\hat{w_1}, \hat{w_2})^2)} \right)
\end{align*}

\paragraph{Linear dependence measure.}
When we assume $w_1$ and $w_2$ follow a multi-variate normal distribution, the GMI measure is only dependent on the covariance $\Sigma$ of the joint distribution $p(w_1, w_2)$
\begin{align*}
    \text{GMI}(w_1,w_2) &= \text{GMI}(\Sigma) = \frac{1}{2} \left(\log\left( \frac{\det\Sigma_{w_1} \det\Sigma_{w_2}}{\det\Sigma}\right) \right)
\end{align*}
Therefore, this measure only considers linear dependencies in the non-collapsed and collapsed form.

\section{Experimental details on encoder training}\label{appendix:encoder_training}
We trained a Resnet-18 encoder \citep{he2016deep} from scratch and added a ReLU and a linear layer to map to the final latent representation $w \in \mathcal{W}$. We set $\lambda_{DC}=0.5$, a batch size of 512 using the Adam optimizer ($\beta_1 = 0.9, \beta_2 = 0.99,\epsilon=10^{-8}$) with a learning rate of 1e-3 for 100 epochs. To avoid overfitting, we used a weight decay of 8e-3. We trained on two 2080 Ti GPUs for 5-6 hours with the PyTorch Lightning framework \citep{falcon2019light}. We chose the best model in terms of total validation loss (early stopping) for evaluation. For t-SNE visualization we used openTSNE \citep{opentsne2019} with an euclidean metric and a perplexity of $n_p$/100, where $n_p$ is the number of data points.

\section{Experimental details on generative model training}\label{appendix:gan_training}
In practice, we trained our GAN models on multiple GPUs. The standard setup in PyTorch Lightning \citep{falcon2019light} for multi GPU training is the Distributed Data Parallel (DDP) strategy. In DDP, each GPU trains with its own data and synchronizes the gradients with all other GPUs. Therefore, each GPU only processes a subset of the whole data batch. However, the batch size is an integral part of the distance correlation estimation (Sec.\,\ref{subsec:disentanglement_loss}). Therefore, we introduced a data ring buffer on every GPU that stores a number of data batches from the past for distance correlation computation.

We trained our model with StyleGAN's style mixing \citep{karras2020analyzing}, which is a regularization technique for disentanglement where images are generated from two different latent vectors that are fed to the generator at different resolution levels. It becomes impractical to train the GAN-inversion in these steps because images generated with style mixing have no clear assignation in the latent space \citep{quiros2021adversarial}. Therefore, we only optimized for GAN-inversion when the generator was not trained with style mixing regularization \citep{karras2020analyzing}. We ended up performing style-mixing regularization only 50\% of the time instead of 90\% as in the original StyleGAN2 implementation. Thus, on average, the GAN-inversion losses are included in every second optimization step.

For the generative model we set $\lambda_{P_L}=2$, $\lambda_{w}=1$, $\lambda_{p}=2$, $\lambda_{C}=0.04$ and $\lambda_{DC}=0.2$. To set $\lambda_{R_1}$ we used a heuristic formula from the original StyleGAN2 implementation which is $\lambda_{R_1}=0.0002 \cdot 256 / (n \cdot g)$ where 256 is the image size, $n$ is the batch size and $g$ is the number of GPUs. The mapping networks had 8 fully connected layers. We trained with a batch size of 56 distributed over 4 GPUs, where on each GPU we had a data ring buffer storing 5 batches (4 batches from the past) for distance correlation computation. For the generator and the discriminator we used individual Adam optimizers, each with a learning rate of 2.5e-3 and hyperparameters $\beta_1 = 0.9, \beta_2 = 0.99,\epsilon=10^{-8}$. We trained each generative model for 200 epochs on four V100 GPUs for 2.5 days with the PyTorch Lightning framework \citep{falcon2019light}. For evaluation, we chose the best model in terms of total validation loss (early stopping). For the t-SNE visualizations we used openTSNE \citep{opentsne2019} with an euclidean metric and a perplexity of $n_p$/100, where $n_p$ is the number of data points.

\section{Experimental details on controllable image generation}\label{appendix:encoder_age_space}
For the age classification model, we trained a Resnet-18 encoder \citep{he2016deep} from scratch,  where we added a ReLU and a linear layer that mapped to the logits $\hat{y} \in \mathbb{R}^3$. We trained the classification model for the 3-class age problem on the reconstructions, since we may have a distribution shift between the real images and the reconstructions. For testing, we swapped age subspaces by first shuffling the test dataset in a deterministic manner and then swapping the age subspaces with a shift of one in each batch. Therefore, each batch element was assigned the age subspace of its predecessor, while the first batch element was assigned the age subspace of the last batch element.

\section{Failure cases}
\begin{figure}[htbp]
    \centering
    \includegraphics[width=\textwidth]{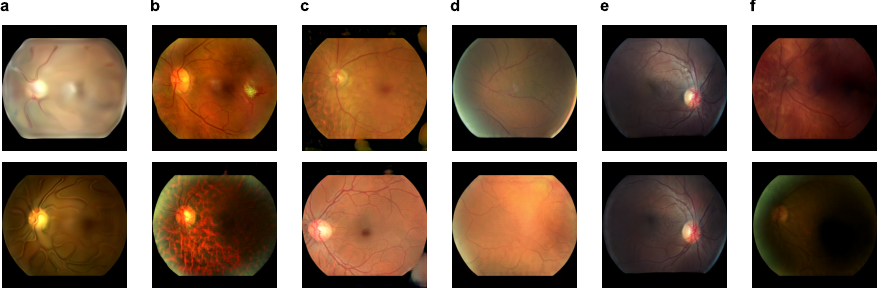}
    \caption{\textbf{Examples of incorrectly generated fundus images.} In \textbf{a}, we show cartoon-like images due to a not yet converged generative model. In addition, training instabilities resulted in unrealistic features in \textbf{b}, non-black background pixels in \textbf{d}, and missing anatomical features in \textbf{d}. Moreover, due to an imperfectly filtered dataset, the optic disc was on the wrong side in \textbf{e}, and some images were of poor quality as in \textbf{f}.}
    \label{fig:appendix-failure-cases}
\end{figure}

We qualitatively investigated the limitation of our generative model by checking for failure cases of a model trained for age-camera-style subspace disentanglement. We randomly sampled latent vectors from a standard normal distribution $z \sim \mathcal{N}(0,I) \in \mathcal{R}^{32}$ over different training epochs. We chose our best model in terms of validation loss. At the beginning of the training, when the generative model had not yet converged, the image features were cartoon-like (Fig.\,\ref{fig:appendix-failure-cases}\,\textbf{a}, epoch 9 and 17). In later stages of training, the fundus images got more realistic, however, we discovered some unrealistic features, such as images that were not perfectly circular (Fig.\,\ref{fig:appendix-failure-cases}\,\textbf{b}, first row, left edges), second optic discs (Fig.\,\ref{fig:appendix-failure-cases}\,\textbf{b}, first row, right center), or extreme colors and shapes (Fig.\,\ref{fig:appendix-failure-cases}\,\textbf{b}, second row). Sometimes images were generated with the non-black pixels in the background (Fig.\,\ref{fig:appendix-failure-cases}\,\textbf{c}) and few images even missed important anatomical structures such as the optic disc and the macula (Fig.\,\ref{fig:appendix-failure-cases}\,\textbf{d}). In some cases, we saw fundus images where the optic disc was on the wrong side (Fig.\,\ref{fig:appendix-failure-cases}\,\textbf{e}), and sometimes the images even looked very similar even though they were generated from different latent spaces (Fig.\,\ref{fig:appendix-failure-cases}\textbf{e}, first versus second row). We even discovered a few blurry images of poor quality (Fig.\,\ref{fig:appendix-failure-cases}\,\textbf{f}).

Unrealistic fundus images (Fig.\,\ref{fig:appendix-failure-cases}\,\textbf{b}) are probably due to a known problem with instability in GAN training due to minimax optimization \citep{becker2022instability}. A second optic disc (Fig.\,\ref{fig:appendix-failure-cases}\,\textbf{b}, first row) or the disc on the wrong side (Fig.\,\ref{fig:appendix-failure-cases}\,\textbf{e}) was probably generated because our data could not be preprocessed perfectly and we had some images with optic discs on the right side in the training data. The same explanation could apply to the blurred images (Fig.\,\ref{fig:appendix-failure-cases}\,\textbf{f}). We have filtered our data set for good quality images with the labels provided by EyePACS. However, there may be label noise in EyePACS and some poor quality images may have ended up in our training data. Insufficient variability in the generated images (Fig.\,\ref{fig:appendix-failure-cases}\,\textbf{e}) can be attributed to mode collapse \citep{becker2022instability}, which is a known GAN failure case, i.e. the generator model collapses and produces only a limited variety of samples. In our case, however, we did not experience a severe case of mode collapse because the models of each epoch provided a wide variety of fundus images overall.

\section{Different image resolutions}
\begin{figure}[htbp]
    \centering
    \includegraphics[width=\linewidth]{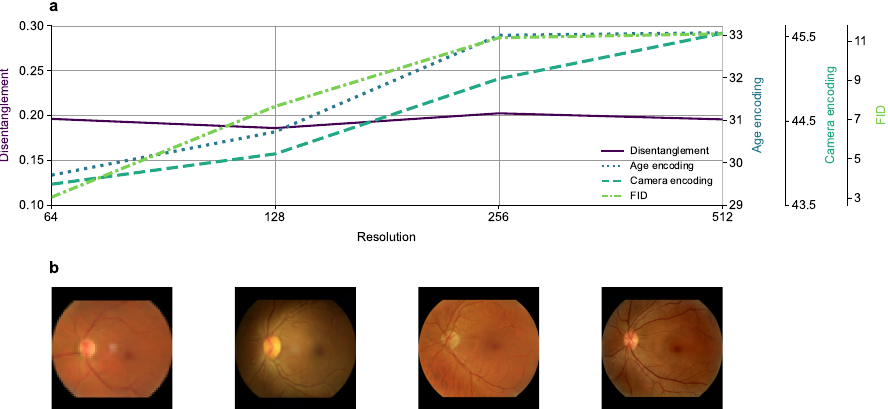}
    \caption{\textbf{Performance comparison of different image resolutions.} In \textbf{a}, we show a selection of measures to evaluate the performance of the GAN models with different resolutions. For panel \textbf{b}, we randomly generated a fundus image with each trained model (resolutions 64, 128, 256, 512; from left to right).}
    \label{fig:appendix-image-resolution}
\end{figure}

We trained generative models with age-camera-style disentanglement with different image resolutions and compared their performance with several metrics: disentanglement measure as the mean of subspace distance correlation values, age and camera encoding, both measured as the above chance level of accuracy, and the FID score as a metric of image quality. Here, we could see that our disentanglement measure was robust over different image resolutions (Fig.\,\ref{fig:appendix-image-resolution},\,\textbf{a}), which was not surprising, since we used the same 32-dimensional latent space for each resolution. Age- and camera-encoding, on the other hand, improves for higher resolution (Fig.\,\ref{fig:appendix-image-resolution},\,\textbf{a}). One possible explanation could be that it is easier to extract age and camera features from images with higher resolution. Finally, the FID score became worse (higher) with increasing resolution (Fig.\,\ref{fig:appendix-image-resolution},\,\textbf{a}). Better (lower) FID scores tend to show greater sample diversity, but the higher resolution images showed qualitatively better image quality at higher resolution (Fig.\,\ref{fig:appendix-image-resolution},\,\textbf{b}). Because we used the same 32-dimensional latent space for each image resolution, larger latent spaces may be required for better FID scores at higher image resolutions.

\section{Multiple subspaces}
\begin{figure}[htbp]
    \centering
    \includegraphics[width=\linewidth]{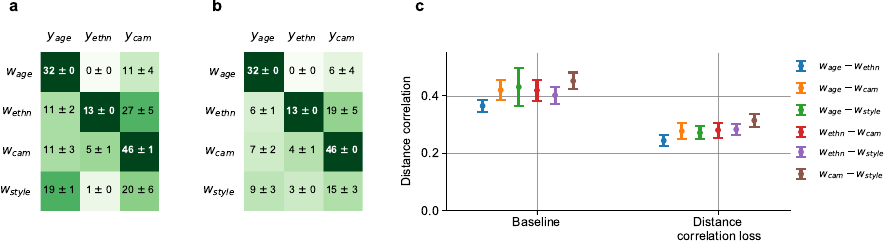}
    \caption{\textbf{Disentanglement performance of a generative model with multiple subspaces.}  We trained generative models on retinal images, featuring four subspaces: $w_\text{age}$ encoding the age class $y_\text{age}$, $w_\text{ethn}$ encoding patient ethnicity $y_\text{ethn}$, $w_\text{cam}$ encoding the camera class $y_\text{cam}$, and $w_\text{style}$ serving as a patient-style subspace. In \textbf{a} and \textbf{b}, we report kNN classifier accuracy improvement over chance level accuracy. The baseline model in \textbf{a} involved training linear classifiers on the first three subspaces. In \textbf{b}, we additionally disentangled the subspaces using our distance correlation loss. Panel \textbf{c} compares the distance correlation measure between subspaces for the two model configurations.}
    \label{fig:appendix-multiple-subspaces}
\end{figure}

To show that our model can handle combinations of multiple attributes, we trained generative models with subspaces for age, ethnicity, and camera encoding simultaneously. Therefore, compared to our previous experiments with 3 subspaces, we scaled up to 4 subspaces, resulting in 6 subspace combinations for distance correlation minimization (Eq.\,\ref{eq:encoder_loss_dcor}). Here we could see that despite training on 4 subspaces, our model still showed comparable performance: the kNN classifier performances on off-diagonals in our confusion matrix of label-subspace combinations for the models with disentanglement loss dropped similarly as in the experiments with only 3 subspaces (Fig.\,\ref{fig:appendix-multiple-subspaces},\,\textbf{a} versus \textbf{b}). In addition, the values of the distance correlation between the individual subspace combinations dropped from an average of 0.42 to 0.28, without major outliers (Fig.\,\ref{fig:appendix-multiple-subspaces},\,\textbf{c}).

\end{appendices}
\end{document}